\PassOptionsToPackage{table}{xcolor}
\documentclass[10pt]{article}
\usepackage{microtype}
\usepackage[preprint]{tmlr}

\usepackage{url}
\usepackage{inconsolata}
\usepackage{graphicx}
\usepackage{amsmath,amssymb,amsfonts}
\usepackage{mathrsfs}
\usepackage{multirow}
\usepackage{booktabs}
\usepackage{textcomp}
\usepackage{enumitem}
\usepackage{hyperref}
\usepackage[hyperpageref]{backref}
\usepackage[font=small]{caption}
\usepackage[bottom]{footmisc}
\usepackage{pifont}
\usepackage[most]{tcolorbox}

\usepackage{algorithm}
\usepackage{algpseudocode}
\usepackage{listings}

\definecolor{AccentColor}{HTML}{916CB4}
\definecolor{AccentColorLight}{HTML}{D8CBE4}
\definecolor{AccentColorSuperLight}{HTML}{EDE7F2}
\definecolor{CiteColor}{HTML}{7340A2}

\usepackage{subfig}
\usepackage{xkcdcolors}
\hypersetup{
    colorlinks=true,
    linkcolor=xkcdOcean,
    urlcolor=xkcdBluish,
    linktoc=all,
    citecolor=CiteColor
}%

\usepackage{amsmath,amsfonts,bm}

\newcommand{\dmodel}{{D_{\text{model}}}}

\def\eqref#1{Eq.~(\ref{#1})}

\def\1{\bm{1}}

\def\rvh{{\bm{h}}}
\def\rvu{{\bm{i}}}
\def\rvj{{\bm{j}}}

\def\rvm{{\bm{m}}}

\def\rvu{{\bm{u}}}
\def\rvv{{\bm{v}}}
\def\rvw{{\bm{w}}}
\def\rvx{{\bm{x}}}
\def\rvy{{\bm{y}}}

\def\vg{{\bm{g}}}

\def\vu{{\bm{u}}}

\def\mU{{\bm{U}}}
\def\mV{{\bm{V}}}
\def\mW{{\bm{W}}}
\def\mX{{\bm{X}}}
\def\mY{{\bm{Y}}}
\def\mZ{{\bm{Z}}}

\DeclareMathAlphabet{\mathsfit}{\encodingdefault}{\sfdefault}{m}{sl}
\SetMathAlphabet{\mathsfit}{bold}{\encodingdefault}{\sfdefault}{bx}{n}

\def\gQ{{\mathcal{Q}}}

\newcommand{\R}{\mathbb{R}}

\usepackage{cleveref}

\usepackage{xspace}
\newcommand{\ourmodel}{Chronos-2\xspace}
\newcommand{\fevbench}{\texttt{fev-bench}\xspace}
\newcommand{\gifteval}{\texttt{GIFT-Eval}\xspace}
\newcommand{\chronosbenchii}{\texttt{Chronos-Bench-II}\xspace}

\hypersetup{pdftitle={Chronos-2: A foundation model for multivariate time series forecasting}, pdfauthor={Abdul Fatir Ansari, Oleksandr Shchur, Jaris Küken, Andreas Auer, Boran Han, Pedro Mercado, Syama Sundar Rangapuram, Huibin Shen, Lorenzo Stella, Xiyuan Zhang, Mononito Goswami, Shubham Kapoor, Danielle C. Maddix, Pablo Guerron, Florian Saupe, Markus Reichstein, Tony Hu, Junming Yin, Nick Erickson, Prateek Mutalik Desai, Hao Wang, Huzefa Rangwala, George Karypis, Yuyang Wang, Michael Bohlke-Schneider}}

\title{\textcolor{AccentColor}{Chronos-2:} A foundation model for multivariate\\ time series forecasting$^{*}$}

\author{\name Abdul Fatir Ansari\textnormal{\textsuperscript{1,\textdagger,\ding{41}}}, Oleksandr Shchur\textnormal{\textsuperscript{1,\textdagger,\ding{41}}}, Jaris Küken\textnormal{\textsuperscript{2,\textdagger,\textdaggerdbl}}, Andreas Auer\textnormal{\textsuperscript{1}}, Boran Han\textnormal{\textsuperscript{1}},\\
Pedro Mercado\textnormal{\textsuperscript{1}}, Syama Sundar Rangapuram\textnormal{\textsuperscript{1}}, Huibin Shen\textnormal{\textsuperscript{1}}, Lorenzo Stella\textnormal{\textsuperscript{1}}, Xiyuan Zhang\textnormal{\textsuperscript{1}},\\
Mononito Goswami\textnormal{\textsuperscript{1}}, Shubham Kapoor\textnormal{\textsuperscript{1}}, Danielle C. Maddix\textnormal{\textsuperscript{1}}, Pablo Guerron\textnormal{\textsuperscript{3,4,\textdaggerdbl}}, Florian Saupe\textnormal{\textsuperscript{1}},\\
Markus Reichstein\textnormal{\textsuperscript{1,6,\textdaggerdbl}}, Tony Hu\textnormal{\textsuperscript{1}}, Junming Yin\textnormal{\textsuperscript{1}}, Nick Erickson\textnormal{\textsuperscript{1}}, Prateek Mutalik Desai\textnormal{\textsuperscript{1}},\\
Hao Wang\textnormal{\textsuperscript{1,5,\textdaggerdbl}}, Huzefa Rangwala\textnormal{\textsuperscript{1}}, George Karypis\textnormal{\textsuperscript{1}}, Yuyang Wang\textnormal{\textsuperscript{1}}, Michael Bohlke-Schneider\textnormal{\textsuperscript{1,\ding{41}}} \\[10pt]
      \addr \textsuperscript{1}Amazon Web Services \textsuperscript{2}University of Freiburg \textsuperscript{3}Amazon \textsuperscript{4}Boston College\\
      \addr \textsuperscript{5}Rutgers University \textsuperscript{6}Max Planck Institute for Biogeochemistry\\
      \addr \textsuperscript{\textdagger}Equal contribution.\\
      \addr \textsuperscript{\ding{41}}Correspondence to: ansarnd@amazon.de, shchuro@amazon.de, bohlkem@amazon.de}

\begin{document}

\begin{tcolorbox}[enhanced, arc=3mm, boxrule=0.6pt, colframe=AccentColor,
    colback=AccentColorSuperLight!18!white, left=10pt, right=10pt, top=8pt, bottom=4pt]
\maketitle

\begin{abstract}
Accurate forecasting with uncertainty quantification is critical in a variety of scientific applications ranging from epidemiology and economics to energy systems and earth sciences. Yet most forecasting models still need to be developed and trained separately for each domain-specific dataset. Time series foundation models (TSFMs) aim to remove this limitation and enable the application of the same model across domains through large-scale pretraining. However, most existing TSFMs treat each time series independently and therefore cannot model interactions among time series and external drivers. Here, we introduce Chronos-2, a foundation model for zero-shot forecasting that models the interactions between multivariate time series and heterogeneous covariates, which commonly arise in scientific applications. The model infers relationships among time series directly from historical context and generates probabilistic forecasts without task-specific treatment. Across various benchmarks, Chronos-2 achieves state-of-the-art accuracy on average by improving skill scores by 5.2 percentage points relative to the strongest baselines and increasing win rates by 10 percentage points. Chronos-2 is also computationally efficient, generating forecasts for 100 time series in under one second on a single mid-range GPU, 50$\times$ and 1000$\times$ faster end-to-end than the best task-specific point and probabilistic forecasters, respectively. In addition, we demonstrate its effectiveness in diverse scientific settings --- including infectious-disease forecasting, macroeconomic modeling, carbon-flux prediction, and runoff simulation --- where the model matches or surpasses specialized methods developed for each domain. These results show that a single pretrained model can capture structure shared across temporal systems. Importantly, our work lays a foundation toward a transition from task-specific forecasting to generalist forecasting systems across scientific and industrial domains.
\end{abstract}
\end{tcolorbox}
{\renewcommand{\thefootnote}{}%
\footnotetext{\hspace{-1.8em}$^{*}$\,This manuscript extends our earlier technical report on Chronos-2~\citep{ansari2025chronos} with scientific case studies, additional baselines, ablation studies, and calibration analyses.}%
\footnotetext{\hspace{-1.8em}$^{\ddagger}$\,Jaris Küken contributed to this work during an internship at AWS. Hao Wang, Pablo Guerron and Markus Reichstein hold concurrent appointments at Amazon and their corresponding institutions, and this manuscript describes work performed at Amazon.}}

\section{Introduction}\label{sec1}

Many natural and human-centric systems evolve as coupled processes. For example, epidemics spread across interconnected populations~\citep{vespignani2020modelling}, economic indicators co-evolve through complex causal relationships~\citep{sims1980macroeconomics}, ecosystems exchange energy and carbon through linked feedback systems~\citep{reichstein2019deep}, and weather events drive river runoff across hydrological basins~\citep{beven2012rainfall}. These phenomena are typically observed as collections of time series, whose joint evolution encodes the underlying dynamics of the system. Accurate forecasting of this temporal data is central to decision-making across disciplines ranging from public health to macroeconomics to energy systems to earth sciences.

Despite decades of progress, forecasting approaches remain largely application-specific due to the heterogeneity of temporal patterns and task characteristics --- including varying granularities, context lengths, missing values, and diverse multivariate interactions, both among targets (the primary variables to be forecast) and with external covariates (auxiliary variables that provide predictive context). Classical statistical models~\citep{hyndman2018forecasting} effectively capture simple temporal structure, but struggle to leverage large-scale datasets or jointly model many related variables. Modern machine learning methods~\citep{salinas2020deepar,lim2021temporal,Nie2023PatchTST} can represent more complex dynamics, yet typically require task-specific training and careful tuning for each dataset. In contrast, recent time series foundation models (TSFMs) enable \emph{zero-shot} forecasting on unseen datasets through large-scale pretraining~\citep{ansari2024chronos, das2024decoder, woo2024unified, cohen2025time}. However, most TSFMs still treat each time series in isolation~\citep{ansari2024chronos, das2024decoder}, which prevents them from capturing the multivariate interactions that characterize real-world systems. While some recent TSFMs have attempted multivariate pretraining~\citep{woo2024unified, ekambaram2024tinytimemixersttms, taga2025timepfn}, none have demonstrated effective zero-shot multivariate forecasting with covariates.

To address the aforementioned limitations, we introduce \ourmodel, a multivariate foundation model that forecasts collections of time series in a zero‑shot manner. Central to our approach is \emph{in-context learning} (ICL): rather than fitting parameters separately for each task, \ourmodel infers multivariate dependencies directly from historical context at inference time.
This capability is enabled by two key design choices: a novel group-attention mechanism and a diverse synthetic multivariate pretraining corpus.
The group-attention mechanism enables information aggregation across arbitrary sets of target time series and heterogeneous covariates, which supports their joint forecasting. The synthetic pretraining corpus exposes the model to diverse instantaneous and lagged multivariate dependencies, encouraging generalization across domains.

We evaluate \ourmodel on comprehensive forecasting benchmarks, where it outperforms (in aggregate) state-of-the-art models, i.e., task-specific statistical and machine learning methods, and existing TSFMs, without any task-specific training. \ourmodel is also 50$\times$ and 1000$\times$ faster end-to-end than the best task-specific point and probabilistic forecasters; it can process 100 time series in 0.8 seconds on a mid-range NVIDIA A10G GPU. Additionally, \ourmodel can be deployed on standard CPU hardware without specialized accelerators due to its lightweight size of 120M parameters. Lastly, we demonstrate the generalization of the model across scientific applications --- infectious‑disease dynamics, macroeconomic forecasting, ecosystem carbon‑flux prediction, and hydrology --- where it produces accurate probabilistic forecasts and often matches or surpasses specialized models used in these domains.

To summarize, our results suggest that the dynamics of complex temporal systems can be inferred directly from contextual observations using a single pretrained model. This work shifts the focus of the forecasting field from dataset‑specific model fitting toward contextual inference over interacting time series.

\section{A unified architecture for multivariate temporal dynamics}\label{sec:framework}

\begin{figure}[htb]
    \centering
    \includegraphics[width=\linewidth]{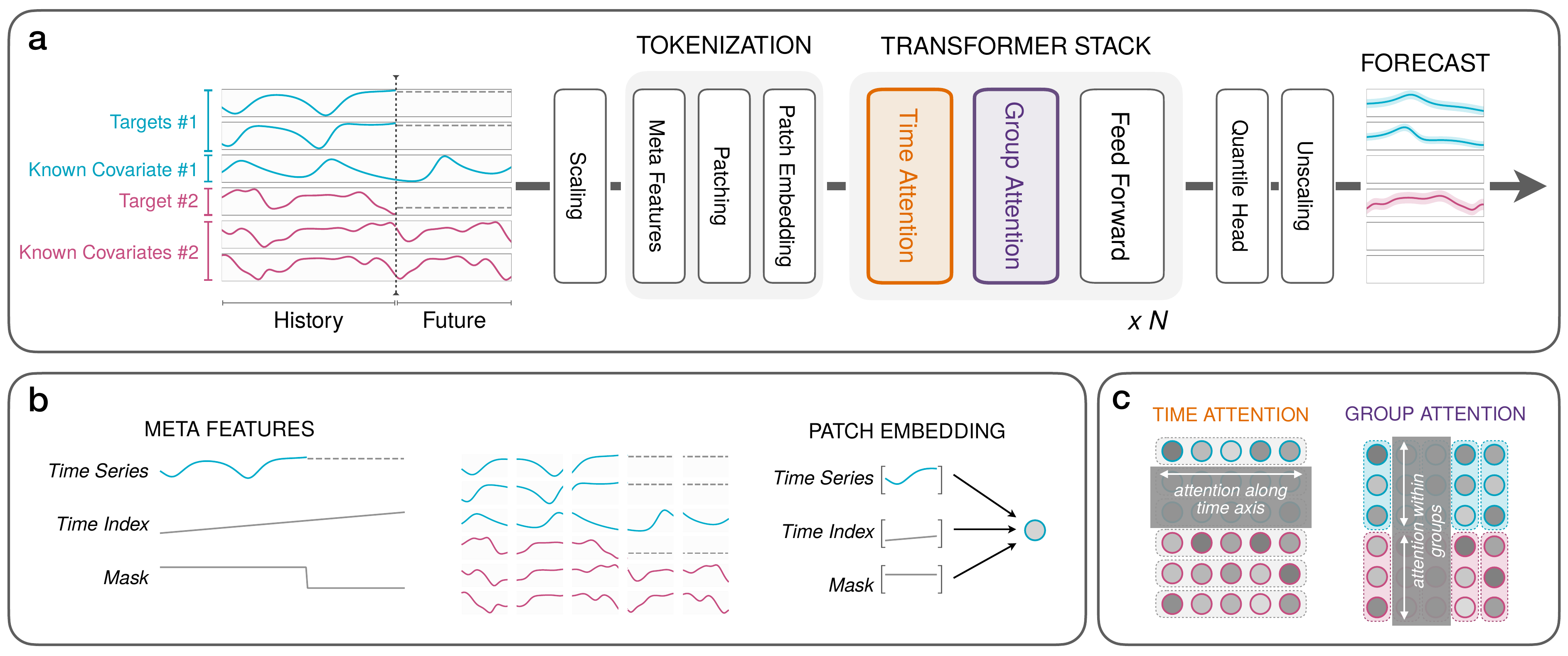}
    \caption{\textbf{The complete \ourmodel pipeline.} (a) \textit{Architecture overview}: Input time series (targets and covariates) are first normalized (scaling) and then tokenized for downstream processing. The Transformer stack alternates between time and group attention layers. The output from the Transformer stack is mapped to quantile predictions through a neural network. The figure illustrates two time series tasks, the first (in {\textcolor[HTML]{00ACCB}{cyan}}) with two targets and one known (future) covariate, and the second (in {\textcolor[HTML]{C64D80}{pink}}) with one target and two known covariates.
    (b) \textit{Tokenization}: Meta features are appended to the time series which indicate the position of an observation (time index) and whether it is missing (mask). Time series and meta features are split into non-overlapping patches and mapped to high-dimensional embeddings via a neural network. (c) \textit{Attention layers}: Each Transformer block alternates between time and group attention layers: the time attention layer aggregates information across patches within a single time series, and the group attention layer aggregates information across all series within a group at each patch index. The groups corresponding to the two time series tasks are highlighted in {\textcolor[HTML]{00ACCB}{cyan}} and {\textcolor[HTML]{C64D80}{pink}}. A group is a flexible notion of relatedness and may correspond to a single time series, multiple series sharing a source or metadata, variates of a multivariate series, or targets along with associated covariates. The group attention layer enables the model to infer relationships between different time series in the group based on the available context at inference time (in-context learning).}
    \label{fig:arch}
\end{figure}

\ourmodel is a unified foundation model for zero-shot forecasting across univariate and multivariate time series problems (\Cref{fig:arch}). The method has four central components: (i) an encoder-only Transformer~\citep{vaswani2017attention} that alternates temporal attention with group attention to infer dependencies among related series; (ii) a flexible inference procedure dictated by group assignments; (iii) a tokenizer that maps targets and covariates into normalized patch sequences; (iv) a pretraining recipe based on heterogeneous forecasting tasks and a quantile-regression objective (see Appendix~\ref{sec:methods} for specific technical details). In~\Cref{sec:data}, we discuss the training corpus, which combines real and synthetic univariate time series with synthetic multivariate systems. 

\ourmodel adopts the Transformer architecture~\citep{vaswani2017attention} --- specifically, a 120M-parameter encoder-only model derived from the T5 architecture~\citep{raffel2020exploring} --- as its core computational building block. Transformers use the attention mechanism to selectively aggregate information based on underlying relationships. \ourmodel employs attention along two dimensions: the time attention layer aggregates information along the temporal axis within each series, and the group attention layer aggregates information across related series at a shared time index. 

 Group attention allows \ourmodel to model complex relationships between groups of time series. Here, a \emph{group} is a flexible notion of relatedness that enables diverse downstream forecasting scenarios. Changing group identities allows the model to seamlessly switch between univariate and multivariate forecasting modes at inference time.

Time series foundation models need the flexibility to handle real-world data formats, e.g., differing scales, covariate types and missing values. To stabilize the heterogeneous scales encountered in practice, \ourmodel represents targets and covariates in a common token space where each input dimension is normalized using historical statistics. Additionally, future values of certain covariates (e.g., planned events or holidays, promotions or weather forecasts) are known in advance and can be useful in several forecasting scenarios. \ourmodel integrates these known covariates by taking both the historical context and the future horizon of each time series as input. We use a binary mask to indicate unknown future values (targets and past-only covariates), which are zero-filled. This masking strategy also accommodates missing values, which are ubiquitous in scientific and industrial forecasting.

\ourmodel splits the time series into non-overlapping patches to improve memory usage and inference speed~\citep{Nie2023PatchTST, das2024decoder, woo2024unified}. The patches are embedded via a neural network into a sequence of high-dimensional vectors for processing. Importantly, \ourmodel decodes multiple forecast patches in a single forward pass rather than relying on expensive, autoregressive step-by-step generation. This patch-based design enables the model to process long contexts and horizons, handling up to 8,192 historical time steps and predicting up to 1,024 future steps at once, while remaining efficient enough to run on consumer-grade GPUs or standard CPU machines.

Lastly, \ourmodel is probabilistic by design to support reliable downstream decision-making. A quantile head outputs a set of quantiles rather than a single point estimate, which enables probabilistic forecasts without imposing a strict parametric form (e.g., a normal distribution) on the predictive distribution. Quantile regression~\citep{koenker1978, wen2017} captures distributions with complex or multimodal shapes. The inclusion of extreme quantiles (0.01 and 0.99) improves the coverage of rare events, which is critical in scientific applications~\citep{camps-valls2025ai}.

\section{Synthesizing diverse multivariate dynamics}
\label{sec:data}

\begin{figure}[htb]
    \centering
    \includegraphics[width=\linewidth]{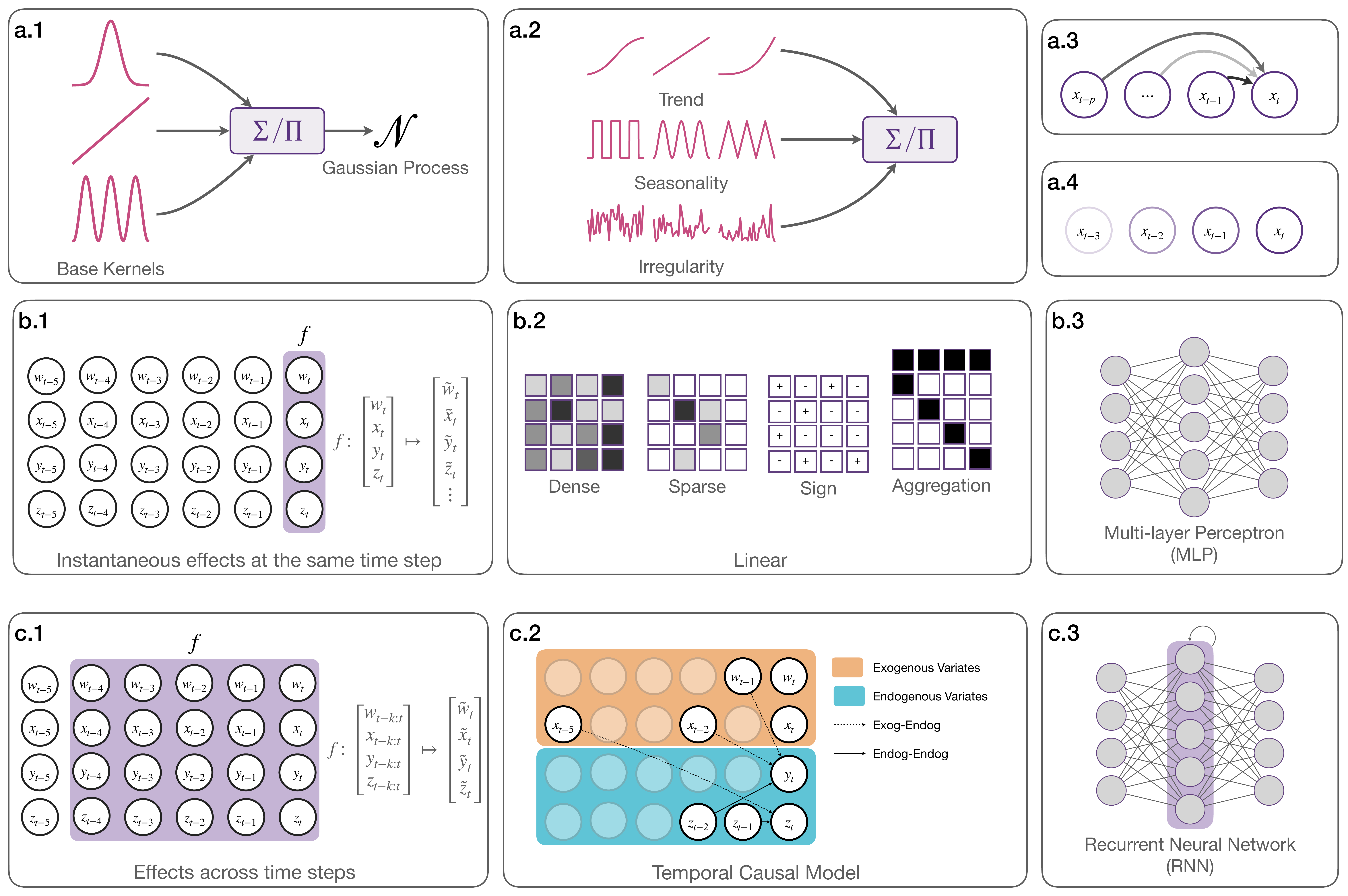}
    \caption{\textbf{Generation of multivariate synthetic training data}. We generate multivariate synthetic time series by applying random transformation functions to time series sampled from univariate generators. \textit{Univariate generators} produce diverse temporal patterns to serve as the building blocks for multivariate data. These include: (a.1) KernelSynth~\citep{ansari2024chronos}, which draws samples from Gaussian processes defined by the random composition of basic kernels (e.g., linear, periodic); (a.2) Trend-Seasonality-Irregularity (TSI), which stochastically combines structural components; and parametric simulations using random (a.3) Autoregressive (AR) and (a.4) Exponential Smoothing (ETS) processes. (b.1) \textit{Instantaneous transformations} impose coupling between variables without time delays. A mixing function transforms univariate samples at each time step. We employ (b.2) random linear operators (e.g., random dense, sparse, sign or aggregation matrices) and (b.3) non-linear mappings via random Multi-Layer Perceptrons (MLPs) to simulate complex static dependencies. (c.1) \textit{Sequential transformations} introduce time-delayed relationships between time series. A random function maps a look-back window of univariate series to multivariate outputs. We generate (c.2) linear lead-lag relationships via a Temporal Causal Model, which samples sparse directed edges across time and variates (simulating Vector Autoregression), while (c.3) non-linear dynamic correlations are introduced via random Recurrent Neural Networks (RNNs).}
    \label{fig:data-fig}
\end{figure}

In addition to the model architecture, the pretraining data plays an important role when developing a foundation model~\citep{touvron2023llama, hollmann2025tabpfn}. 
While recent efforts have expanded the availability of large-scale time series datasets, they remain primarily univariate~\citep{woo2024unified, ansari2024chronos}. To overcome this limitation, we develop a scalable pipeline for generating synthetic multivariate data that covers a comprehensive space of temporal patterns and interactions. We train \ourmodel on this dataset to equip it with the capability to understand multivariate interactions through ICL.

Our generation process mimics the complexity of real-world systems by hierarchically composing simple signals into complex interactions. We begin by sampling univariate time series from a diverse suite of generators (\Cref{fig:data-fig}a), ranging from structural models that capture trends and seasonality~\citep{bahrpeyma2021methodology} to stochastic processes, e.g., exponential smoothing~\citep{hyndman2008forecasting} and Gaussian processes~\citep{ansari2024chronos}. These act as the fundamental building blocks, which provide a rich variety of univariate temporal dynamics.

We construct multivariate time series by coupling these univariate time series through two distinct interaction mechanisms: instantaneous and sequential.

Instantaneous interactions commonly occur in real-world applications, e.g., multiple sensors in a machine responding to a load change or retail sales across different store locations responding to the same holiday or weather event. These interactions introduce relationships among time series where there is no time delay. To capture these dynamics, we apply transformations across the univariate series at the same time step (\Cref{fig:data-fig}b). We model linear dependencies using mixing matrices to generate dense or sparse correlations. Non-linear interactions are synthesized via multi-layer perceptrons (MLPs) that map concurrent observations to a new multivariate space, creating non-linear relationships between variables at each time step.

Sequential interactions also occur in real-world systems, e.g., infection rates leading to hospitalization spikes or marketing spending preceding sales growth. These interactions often exhibit time-delayed relationships, where an event in one variable triggers a response in another after a specific duration. To capture these dynamics, we introduce lagged dependencies using two distinct mechanisms (\Cref{fig:data-fig}c). For linear temporal relationships, we synthesize Vector Autoregressive (VAR) processes~\citep{lutkepohl2013introduction} based on randomly sampled causal graphs by explicitly defining which variables drive others over time. For non-linear temporal dynamics, we apply Recurrent Neural Networks (RNNs)~\citep{hochreiter1997long, goodfellow2016deep} to the univariate history to create systems where the future evolution depends on complex, history-dependent interactions between variables that cannot be captured by simple linear lags.

In summary, we scale up pretraining data and introduce multivariate relationships through multivariate synthetic data generation. We train~\ourmodel on this corpus to enable it to infer the relationship between time series from the available context and react to them at inference time.

\begin{figure}[htb]
    \centering
    \includegraphics[width=\linewidth]{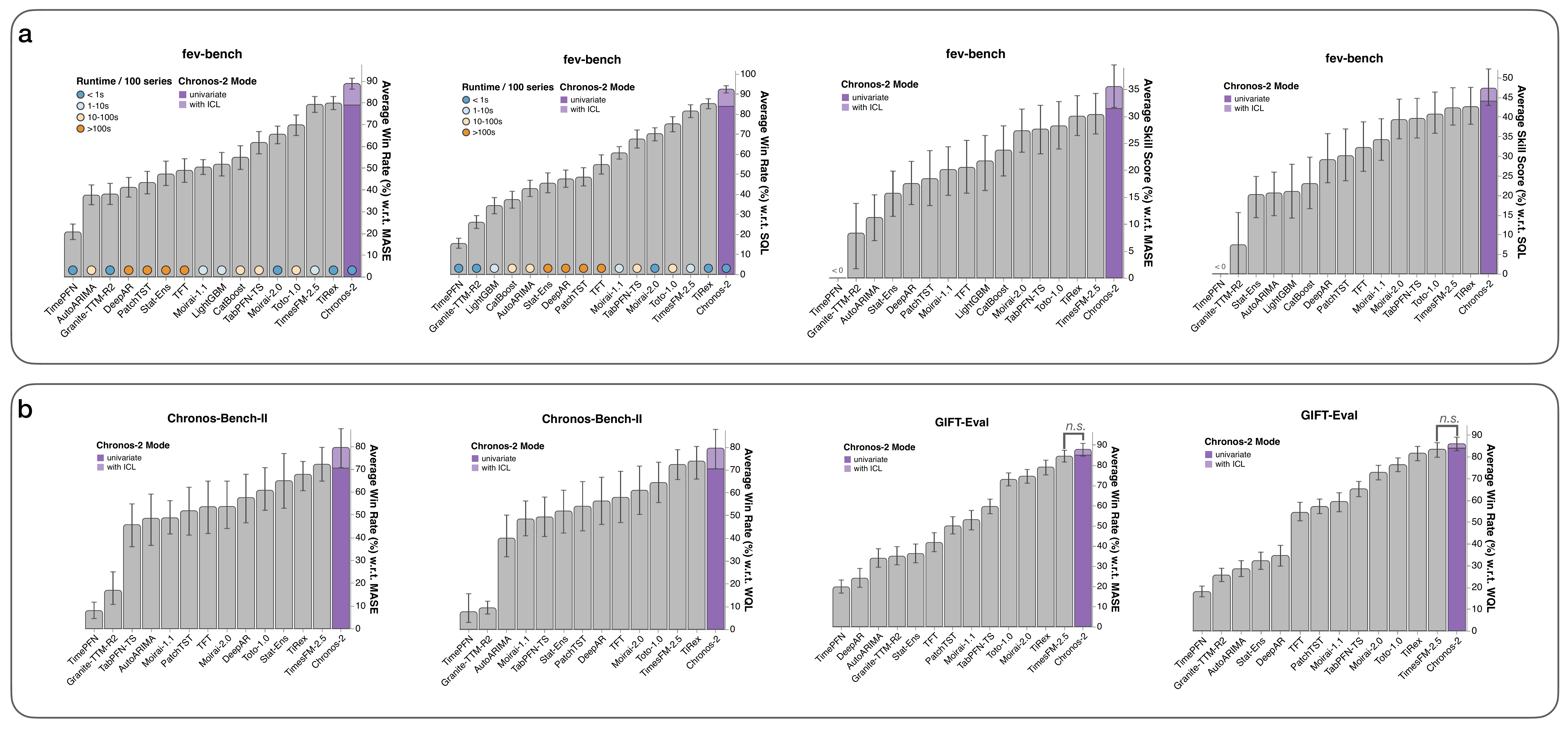}
    \caption{\textbf{Evaluation of \ourmodel on time series forecasting benchmarks} (\textit{best viewed in color on a screen}). Performance is evaluated using the Mean Absolute Scaled Error (MASE) and Scaled/Weighted Quantile Loss (SQL/WQL) for point and probabilistic forecasts, respectively. The Win Rate measures the proportion of pairwise comparisons where a model outperforms competitors, while the Skill Score reflects the average percentage improvement over a Seasonal Naive baseline. Dark purple bars denote \ourmodel in univariate mode (group attention disabled); light purple segments highlight the performance gain attributed to in-context learning (ICL) via group attention. Error bars indicate 95\% confidence intervals obtained through bootstrapping. (a) Results on \fevbench ($n=100$ tasks), a rigorous benchmark where 68 tasks involve multivariate dependencies. \ourmodel outperforms all baselines, including statistical ensembles, tree-based methods, task-specific deep learning models, and existing foundation models, while maintaining fast inference speed; Wilcoxon signed-rank test $p < 10^{-6}$ against all baselines (see \Cref{tab:fev-results-mase,tab:fev-results-sql} in the Appendix for results in tabular format and \Cref{tab:fev-bench-pval} for the $p$-values). (b) Results on \chronosbenchii ($n=27$ tasks) and \gifteval ($n=97$ tasks), which focus primarily on univariate forecasting. \ourmodel outperforms or ties state-of-the-art methods; notably, enabling ICL yields performance improvements even in these univariate settings by retrieving patterns across related series. \textit{n.s.} indicates that the performance difference relative to \ourmodel is not statistically significant ($p \geq 0.05$) based on the Wilcoxon signed-rank test; see \Cref{tab:zs-mase,tab:zs-wql,tab:gift-eval-mase,tab:gift-eval-wql} in the Appendix for results in tabular format and \Cref{tab:zs-pval,tab:gift-pval} for the $p$-values.}
    \label{fig:main-results}
\end{figure}

\section{Zero-shot generalization across benchmarks}\label{sec:main-results}

We assess the performance of \ourmodel on \fevbench~\citep{shchur2025fev}, a comprehensive benchmark comprising 100 forecasting tasks designed to evaluate time series models in a standardized setting.
The \fevbench benchmark contains tasks spanning domains such as energy, retail, cloud computing, economics, healthcare, and mobility.
It emphasizes real-world heterogeneity with 68 of the tasks involving multivariate dependencies.
These tasks require the model to handle multiple co-evolving target variables, past-only and known (future) covariates, and diverse data types (numeric, boolean, and categorical), often in the presence of missing values.
To complement this multivariate focus, we also evaluate \ourmodel on \chronosbenchii~\citep{ansari2024chronos} and \gifteval~\citep{aksu2024gifteval}, which are two established benchmarks comprising 27 and 97 primarily univariate tasks, respectively.

We benchmark \ourmodel against a diverse spectrum of forecasting methods, including statistical models (AutoETS, AutoARIMA, AutoTheta, and their ensembles)~\citep{hyndman2018forecasting, petropoulos2020simple}, tree-based methods (LightGBM~\citep{ke2017lightgbm}, CatBoost~\citep{prokhorenkova2018catboost}), task-specific deep learning models (DeepAR~\citep{salinas2020deepar}, TFT~\citep{lim2021temporal}, PatchTST~\citep{Nie2023PatchTST}), and state-of-the-art TSFMs (TiRex~\citep{auer2026tirex}, TimesFM-2.5~\citep{das2024decoder}, Toto-1.0~\citep{cohen2025time}, Moirai-1.1~\citep{woo2024unified}, Moirai-2.0~\citep{liu2025moirai}, Granite-TTM-R2~\citep{ekambaram2024tinytimemixersttms}, TimePFN~\citep{taga2025timepfn} and TabPFN-TS~\citep{hollmann2025tabpfn, hoo2025tables}).
\ourmodel and other foundation models are pretrained and used directly without fitting on individual tasks in the benchmarks (zero-shot inference).

We evaluate the performance using Mean Absolute Scaled Error (MASE) for point forecasting and Scaled/Weighted Quantile Loss (SQL/WQL) for probabilistic forecasting. We report two complementary metrics: the win rate, representing the proportion of pairwise comparisons where a model outperforms its competitors, and the skill score, which reflects the average percentage improvement over a baseline (Seasonal Naive).
Additionally, we report the improvability~\citep{erickson2026tabarena}, which quantifies the gap between the accuracy of a model and the best (in hindsight) model per task.
A lower average improvability indicates that a model is consistently close to the best available, even when it is not the top performer on every individual task.

\Cref{fig:main-results}(a) illustrates the strong performance of \ourmodel on \fevbench. 
The model establishes a new state-of-the-art in average accuracy by achieving an absolute MASE (SQL) win rate increment of 41.6 (46.9) percentage points over the best statistical model (Statistical Ensemble), 33.9 (55.1) over the best tree-based method (CatBoost), 39.9 (37.6) over the leading deep learning model (TFT), and 9 (7.2) over the strongest TSFM baseline (TiRex). 
On MASE (SQL) skill scores, \ourmodel yields improvements of 11.8 (15.1) percentage points over the best task-specific model (CatBoost/TFT) and 5.2 (4.7) over the strongest TSFMs (TimesFM-2.5/TiRex).
These results also demonstrate the gains attributable to in-context learning: enabling group attention yields absolute improvements of 10.7 percentage points in the win rate and 4.1 in the skill score compared to univariate inference.
In addition to the best skill score and win rate, \ourmodel also achieves the lowest MASE (SQL) improvability of 4.5\% (3.6\%) compared to the next best model (TimesFM-2.5/TiRex) at 10.5\% (10.4\%); see \Cref{tab:fev-results-mase,tab:fev-results-sql} in the Appendix. This means that even in tasks where \ourmodel is not the top performer, the margin to the best available model is small, making it the most reliable single-model choice across diverse tasks.

Notably, \ourmodel achieves this accuracy while maintaining high throughput, with an average runtime of less than 1 second per 100 time series on an A10G GPU.
Unlike task-specific models, \ourmodel does not need to be trained for individual tasks and only incurs inference cost. This makes it 50$\times$ and 1000$\times$ faster than the best task-specific point (CatBoost) and probabilistic forecasters (TFT) in terms of end-to-end runtime, which includes training and tuning time for task-specific models; see \Cref{tab:fev-results-mase,tab:fev-results-sql} in the Appendix for per-model runtimes.

Furthermore, we demonstrate that in addition to its multivariate capabilities, \ourmodel is also effective at univariate forecasting.
\Cref{fig:main-results}(b) and \Cref{tab:zs-mase,tab:zs-wql,tab:gift-eval-mase,tab:gift-eval-wql} (Appendix) show that \ourmodel leads on both \chronosbenchii and \gifteval~in terms of win rate, skill score, and improvability.
These benchmarks stress distinct capabilities: \chronosbenchii focuses on short-history tasks, and \gifteval focuses on long-context and long-horizon prediction. 
Interestingly, enabling in-context learning yields improvements even on these univariate tasks. 
This effect is particularly pronounced on \chronosbenchii. These results suggest that when historical data is scarce, the group attention mechanism allows the model to ``cross-learn'' and stabilize predictions by retrieving relevant patterns from other time series.

Lastly, we assess the calibration of probabilistic forecasts generated by \ourmodel against models that support extreme quantiles. We compute the Weighted Interval Score (WIS), a proper scoring rule for evaluating the sharpness and calibration of predictive distributions~\citep{gneiting2007strictly}, on three prediction intervals: 90\% $[0.05, 0.95]$, 95\% $[0.025, 0.975]$, and 98\% $[0.01, 0.99]$. In addition, we report the Mean Absolute Coverage Error, which directly measures the discrepancy between nominal and empirical coverage. \Cref{fig:calibration} (Appendix) shows that \ourmodel achieves the best calibration on 23 out of 24 metric-interval pairs on the \fevbench, \chronosbenchii, and \gifteval benchmarks. 

\section{Qualitative analysis on in-context learning}\label{sec:qualitative}

\begin{figure}[ht]
    \centering
    \includegraphics[width=\linewidth]{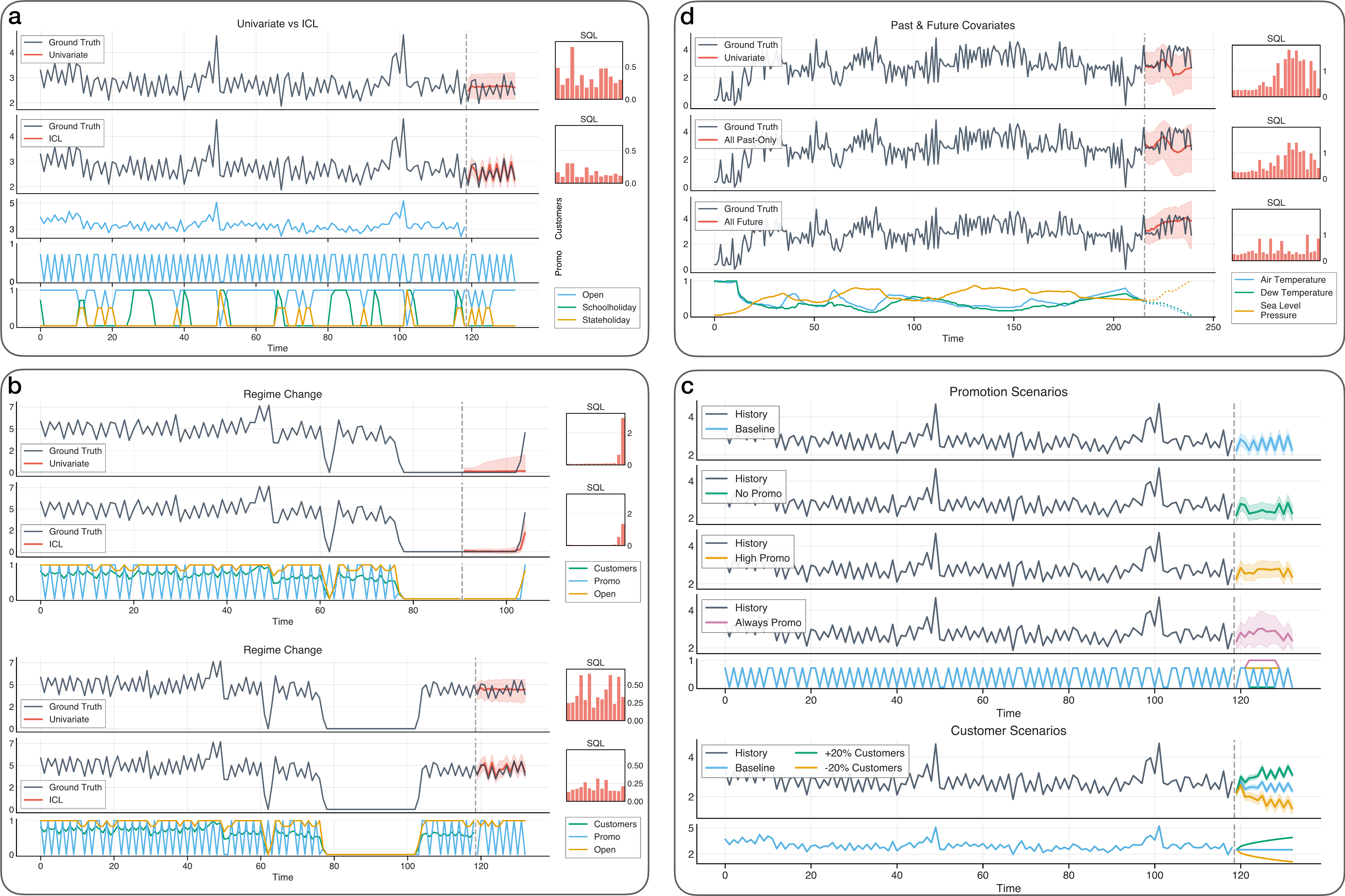}
    \caption{\textbf{Qualitative analysis of in-context learning} (\textit{best viewed in color on a screen}).
    Probabilistic forecasts show the $10\%$ to $90\%$ quantile interval; bar charts show SQL forecast errors per time step.
    (a) Retail sales forecasting: univariate inference versus in-context learning with covariates including promotions, customer traffic, and holiday indicators. Covariates are shown in the lower panels.
    (b) Regime change: store closure at different positions relative to the forecast horizon. The top and bottom rows show the same series with different forecast origins, so that the closure falls within the forecast horizon (top) or entirely within the historical context (bottom).
    (c) Scenario analysis with simulated future covariates: promotion intensity (top four panels) and customer count trajectories (bottom panel) varied across hypothetical trajectories. 
    (d) Energy load forecasting: univariate, past-only covariates, and full covariates (air temperature, dew temperature, sea-level pressure). The three rows show progressively more covariate information available to the model.}
    \label{fig:quali-main-results}
\end{figure}

Here, we aim to understand the mechanisms driving the strong performance on the quantitative benchmarks, which requires taking a deeper look at how the model utilizes context.
In particular, we present a qualitative analysis of scenarios from the retail and energy domains to demonstrate how in-context learning allows the model to understand multivariate interactions, respond meaningfully to counterfactual scenarios, and resolve ambiguity during regime shifts.

In several forecasting tasks, the target variable is heavily modulated by external factors. \Cref{fig:quali-main-results}(a) compares \ourmodel in univariate mode against its default in-context learning-based multivariate mode on a retail sales task from \fevbench.
While the univariate model captures the level of the time series, it fails to anticipate fluctuations tied to external drivers such as promotions and holidays.
By conditioning on these covariates via in-context learning, the multivariate model aligns closely with the ground truth, which results in sharpened prediction intervals and reduced error metrics. 

A critical challenge in forecasting is handling regime changes --- abrupt transitions between distinct behavioral states, such as store closures or system failures~\citep{quandt1958regimes, hamilton1989new}.
Observable covariates often carry information about the latent regime state, which enables covariate-driven regime identification~\citep{filardo1994business, auer2023conformal}. \Cref{fig:quali-main-results}(b) illustrates how \ourmodel exploits this principle: observed covariates allow it to identify regimes that are invisible to a univariate model.
In scenarios where a store closes for an extended period, a univariate model is unaware of the cause of the zero-sales period, which results in inflated uncertainty. 
In contrast, \ourmodel leverages the ``Open'' covariate to disambiguate the context.
The group attention mechanism correctly identifies the closure regime, maintains tight confidence intervals during the inactive period, and anticipates the reopening at the boundary.
This demonstrates the ability of \ourmodel to leverage covariates to switch effectively between dynamic modes.

In addition to passive conditioning, we can also probe the model's understanding of the dynamics by feeding it hypothetical future covariate trajectories. \Cref{fig:quali-main-results}(c) visualizes this by modifying the promotion schedule for a retail sample.
When the promotion activity is set to zero (no promo), the forecast settles to the baseline lower bound; when the promotion is set to high, the forecast shifts to the upper bound.
Notably, when tested with an always-on promotion schedule --- a scenario never seen in the history --- the model predicts high sales but significantly widens its uncertainty bands.
The model's behavior reflects its epistemic uncertainty when forced to extrapolate to out-of-distribution conditions.

\ourmodel can also use past-only drivers, such as customer counts, to simulate different scenarios where we do not know the future values of the covariate. While typically unavailable at forecast time, we can provide the model with synthetic future trajectories (stable average, 20\% increased and 20\% decreased) to simulate potential business scenarios.
The model produces coherent forecasts, where sales volume tracks the simulated footfall intensity.
This indicates that \ourmodel has internalized the structural coupling between footfall and sales, which allows it to function as a forecaster and a simulator for strategic planning.

Finally, real-world deployment often involves scenarios where future values of certain covariates are unavailable.
\Cref{fig:quali-main-results}(d) analyzes an electricity load forecasting task under varying information availability.
We see that while known (future) covariates provide the highest accuracy, \ourmodel extracts significant signal even from past-only covariates, which results in reduced error metrics compared to the univariate baseline.
This shows the model can leverage historical covariate information to inform future dynamics of the target variable, yielding improvements over univariate forecasting.

\section{Case studies}\label{sec:case-studies}

In this section, we apply \ourmodel to various scientific problems --- infectious disease dynamics, consumer price index forecasting, carbon flux prediction, and rainfall-runoff simulation --- and demonstrate its capabilities on hierarchical forecasting. For each case study, we adopt the rigorous evaluation protocols of the respective discipline. In particular, we compare \ourmodel directly against the specialized models and operational ensembles typically employed by domain experts. In addition, we report performance using the standard metrics of each field --- e.g., the Weighted Interval Score in epidemiology or the Nash–Sutcliffe Efficiency in hydrology --- to ensure our results are directly interpretable within the context of the domain-specific literature.

\subsection{Epidemiology: Infectious disease dynamics}

Accurate forecasting of acute respiratory infection (ARI) and influenza-like illness (ILI) is fundamental to public health preparedness, guiding resource allocation and intervention strategies during peak transmission periods~\citep{viboud2019future}. We leveraged data from the ECDC RespiCast Hub~\citep{ecdc_respicast}, a collaborative platform that aggregates predictions from independent modeling teams to support decision-making at the European Centre for Disease Prevention and Control.

We benchmarked the zero-shot performance of \ourmodel against the RespiCast Hub-Ensemble~\citep{ecdc_respicast}, an established operational benchmark that synthesizes outputs from diverse epidemiological approaches ranging from mechanistic compartmental models (SEIR) to statistical baselines. Following the official evaluation protocol, we employed the relative Weighted Interval Score (WIS) to assess the quality of probabilistic forecasts~\citep{bracher2021evaluating}. This comparison is inherently demanding, as the Hub-Ensemble benefits from the ``wisdom of the crowd'' and creates a consensus from multiple domain-specific and statistical signals~\citep{reich2019collaborative}.

\begin{figure}[htb]
    \centering
    \includegraphics[width=\textwidth]{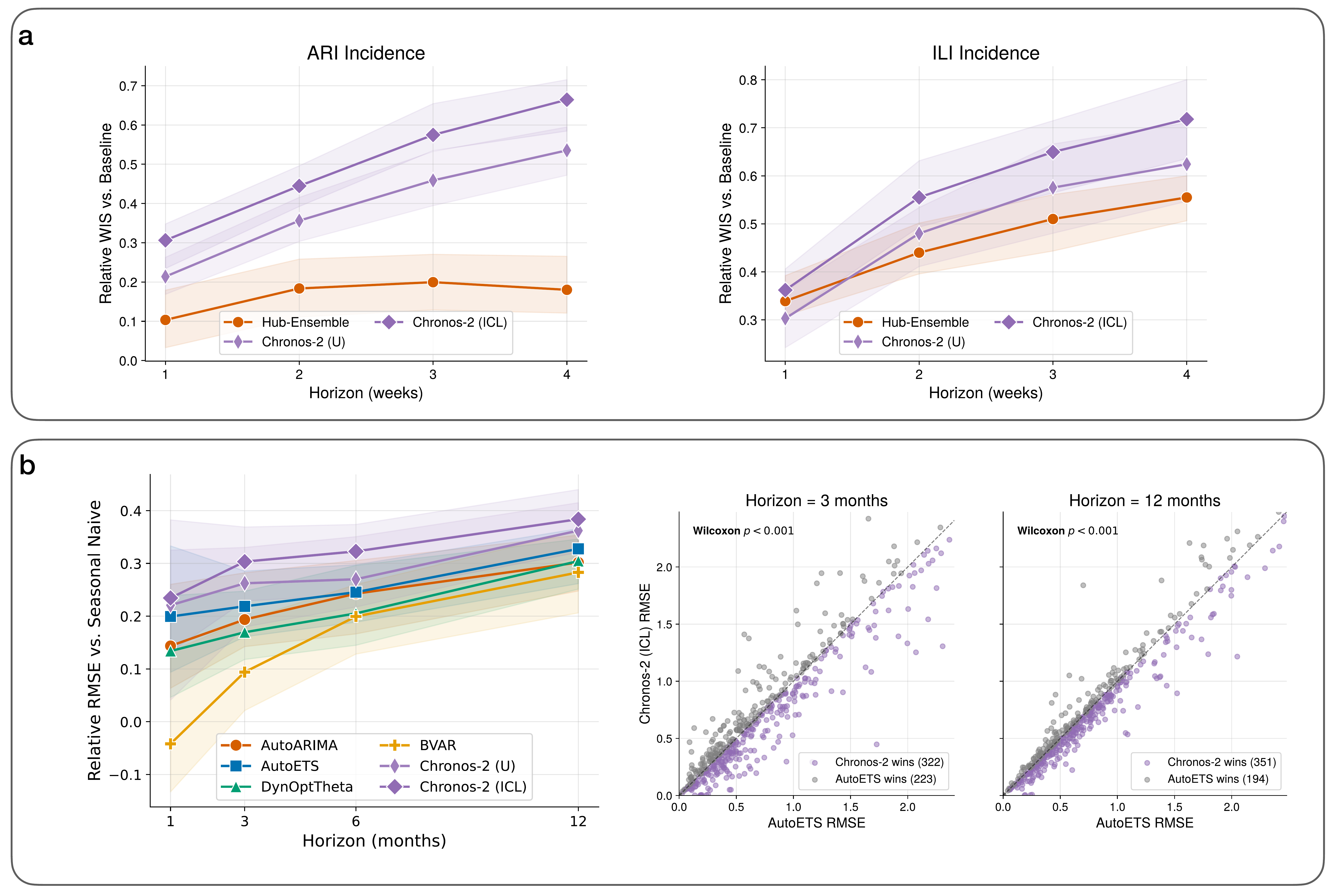}
    \caption{\textbf{Zero-shot forecasting of complex human-centric systems.} (a)
     Forecasting incidence of Acute Respiratory Infection (ARI) and Influenza-Like Illness (ILI) across 31 European countries. Performance is evaluated using the Weighted Interval Score (WIS) relative to a simple quantile baseline provided by RespiCast~\citep{ecdc_respicast}, calculated as $\log_2(\mathrm{WIS}_{\text{baseline}}/\mathrm{WIS}_{\text{model}})$; values greater than 0 indicate improvement over the baseline. Plots show the median relative WIS (solid lines) and 95\% confidence intervals (shaded bands) over 67 rolling evaluation windows (Oct 2024–Jan 2026). In univariate mode (U), \ourmodel is competitive at short horizons (1 week) against the operational RespiCast Hub-Ensemble (a consensus of mechanistic and statistical models). At longer horizons ($>1$ week), \ourmodel significantly outperforms the ensemble, with in-context learning yielding further gains by exploiting cross-country correlations (Wilcoxon signed-rank test $p<0.001$). (b) Joint forecasting of German Consumer Price Index (CPI) components (2021–2025). Performance is measured by the Root Mean Squared Error (RMSE) relative to a Seasonal Naive baseline, $\log_2(\mathrm{RMSE}_{\text{baseline}}/\mathrm{RMSE}_{\text{model}})$. While univariate inference (U) performs comparably to robust statistical baselines, enabling in-context learning significantly improves accuracy by leveraging shared structural patterns within CPI groups. Pairwise comparisons (right) confirm that \ourmodel outperforms the strongest baseline (AutoETS) on the majority of item-window pairs (Wilcoxon signed-rank test $p<0.001$ for both 3-month and 12-month horizons).} 
    \label{fig:epi-econ}
\end{figure}

\ourmodel achieves competitive short-term forecasting performance solely by analyzing historical incidence trajectories (\Cref{fig:epi-econ}(a)).
This suggests that it effectively captures the statistical patterns characteristic of viral transmission without explicit knowledge of pathogen biology or social mixing patterns.
Importantly, \ourmodel surpasses the performance of the Hub-Ensemble at longer forecast horizons, especially in the case of ARI incidence.
The in-context learning mechanism yields further gains here: by jointly predicting incidence rates across European nations, the model leverages cross-country correlations, thereby refining the forecast for the entire continent. These results suggest that general-purpose foundation models can complement specialized epidemiological modeling. Rather than replacing mechanistic approaches, \ourmodel offers a rapidly deployable, high-quality option that can enhance existing ensembles, particularly in scenarios where specialized modeling teams are scarce or require significant lead time for calibration.

\subsection{Macroeconomics: Consumer price index forecasting}

Forecasting the components of the Consumer Price Index (CPI) is fundamental to the conduct of monetary policy~\citep{stockwatson2016}. Challenges in macroeconomic forecasting include lower frequency (monthly) data, limited historical samples, and significant signal-to-noise challenges. In this regime, complex machine learning models frequently fail to generalize, often underperforming simpler statistical baselines~\citep{carriero2025macroeconomic}. Methods such as Bayesian VAR (BVAR)~\citep{giannone2015prior}, ARIMA and ETS are often difficult to surpass, as they capture trends without overfitting to short-term volatility~\citep{hyndman2018forecasting}.

Here, we tackled the challenging problem of jointly forecasting the granular components of the CPI, where existing literature has typically focused on predicting headline inflation as a single aggregate. We evaluated \ourmodel on the German CPI, decomposed into 110 product groups at Level 4 of the Classification of Individual Consumption According to Purpose (COICOP) standard --- a granularity sufficient to isolate sector-specific drivers while smoothing idiosyncratic item-level noise.

The challenge of achieving performance gains in this regime is evident in \Cref{fig:epi-econ}(b), which illustrates the tight performance margins among methods.
\ourmodel in univariate mode performs comparably to robust statistical baselines (AutoARIMA, AutoETS, and DynOptTheta). In-context learning significantly improves the model's capability.
Rather than processing the 110 components individually, ICL jointly forecasts items within each Level 2 COICOP group (e.g., ``Food and non-alcoholic beverages''), exploiting natural clusters of economically related products.
As shown in the pairwise comparisons, \ourmodel (ICL) consistently outperforms AutoARIMA and AutoETS, achieving significantly lower errors on the majority of items (Wilcoxon signed-rank test $p<0.001$) at both 3-month and 12-month horizons. These results show that foundation models yield gains even in regimes traditionally dominated by classical statistical models.

\subsection{Earth Science: Carbon flux prediction}

\begin{figure}[htb]
    \centering
    \includegraphics[width=\textwidth]{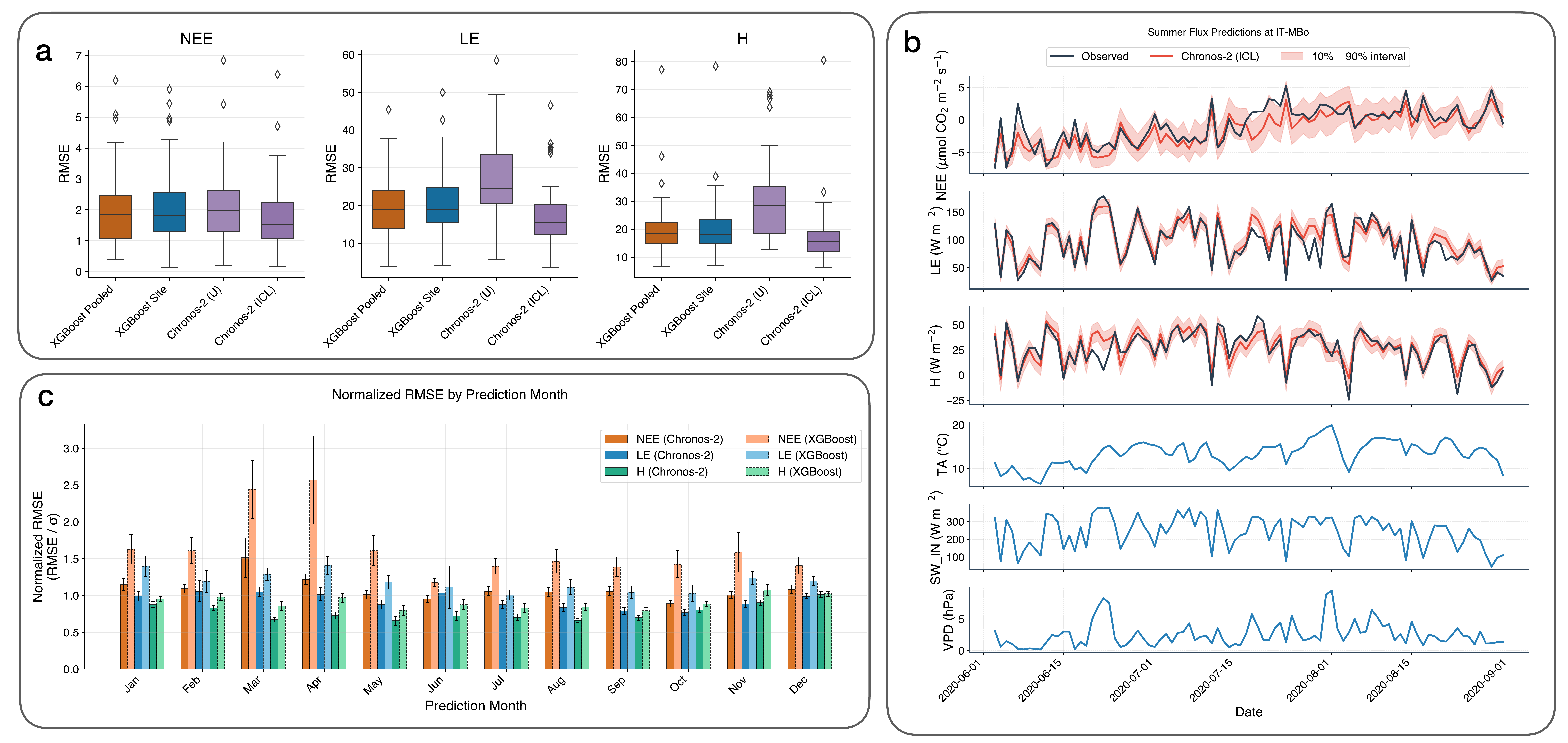}
    \caption{\textbf{Ecosystem carbon and energy flux prediction with \ourmodel.} We evaluate the performance of \ourmodel on daily eddy-covariance targets from the FLUXNET collection \citep{pastorello2020fluxnet2015}. (a) Comparison of the distribution of site-level RMSE across 65 sites for net ecosystem exchange (NEE), latent heat flux (LE), and sensible heat flux (H) over the June--August 2020 evaluation window between supervised regression baselines (XGBoost, trained either as a single pooled model across sites or as site-specific models) and \ourmodel in zero-shot mode. Box plots show the median (center line), interquartile range (box limits), and 1.5$\times$ interquartile range (whiskers); points denote outliers. (b) Representative summer time series at the Monte Bondone site in Italy (IT-MBo), showing observed fluxes (black) and \ourmodel predictions from the meteorological covariates with 10--90\% prediction intervals (red shaded); the corresponding covariates (TA, SW\_IN and VPD) are shown below. (c) Rolling one-month-ahead forecasting over the full year 2020. For each evaluation month, models condition on the preceding 12 months of context and predict the subsequent month (step size: 1 month). Bars report normalized RMSE (RMSE divided by the target standard deviation) stratified by prediction month for \ourmodel and XGBoost, highlighting seasonal variations in predictability across NEE, LE, and H.} 
    \label{fig:carbon}
\end{figure}

Eddy-covariance networks provide the observational backbone for understanding land–atmosphere feedbacks and the terrestrial carbon cycle. We evaluated \ourmodel on the prediction of Net Ecosystem Exchange (NEE) and turbulent energy fluxes --- Latent Heat (LE) and Sensible Heat (H) --- using the standardized FLUXNET collection \citep{pastorello2020fluxnet2015}. While traditional methodologies focus on gap-filling (imputing missing historical data) \citep{fu2021gapfill}, we address the problem of forecasting fluxes based on meteorological drivers. These forward-looking estimates have broad operational utility, ranging from optimizing irrigation in precision agriculture to monitoring natural vegetation stress under increasing atmospheric aridity \citep{doi:10.1126/sciadv.aax1396}. 

We structured this problem as a daily multivariate forecasting task, using air temperature (TA), incoming shortwave radiation (SW\_IN), and vapour pressure deficit (VPD) as covariates. 
We focused on the boreal summer (June–August 2020), the peak period of biological activity and surface-atmosphere exchange. \ourmodel operates entirely zero-shot, whereas conventional baselines are fully supervised, trained on all available historical observations prior to the test period. As the supervised baseline, we selected XGBoost \citep{chen2016xgboost} which is a standard in flux modeling~\citep{fu2021gapfill}. To reflect common modeling choices in heterogeneous multi-site datasets, we considered both a pooled XGBoost model (trained across all sites) and site-specific variants.

\Cref{fig:carbon}(a) shows that \ourmodel with covariates (ICL) achieves the lowest RMSE distributions across NEE, LE, and H (Wilcoxon signed-rank test $p < 0.01$) without requiring any domain-specific training.
ICL clearly outperforms \ourmodel in univariate mode (U), implying that the model effectively leverages meteorological forcing rather than relying on only autoregression. Qualitatively, \ourmodel demonstrates an understanding of physical coupling between meteorological forcing and surface exchange; for instance, it correctly predicts increases in latent heat flux during periods of high shortwave radiation, which is consistent with the fundamental principles of surface energy balance where evapotranspiration is constrained by available energy \citep{moorhead2017sls_lysimeter_et}. Lastly, error stratification by month (Figure \ref{fig:carbon}c) reveals consistent seasonal predictability without requiring training or calibration compared to its task-specific counterpart. Collectively, these results demonstrate that a generalist foundation model can generate physically meaningful and uncertainty-aware ecosystem forecasts that outperform strong supervised regressors.

\subsection{Hydrology: Rainfall-runoff simulation}

\begin{figure}[htb]
    \centering
    \includegraphics[width=\textwidth]{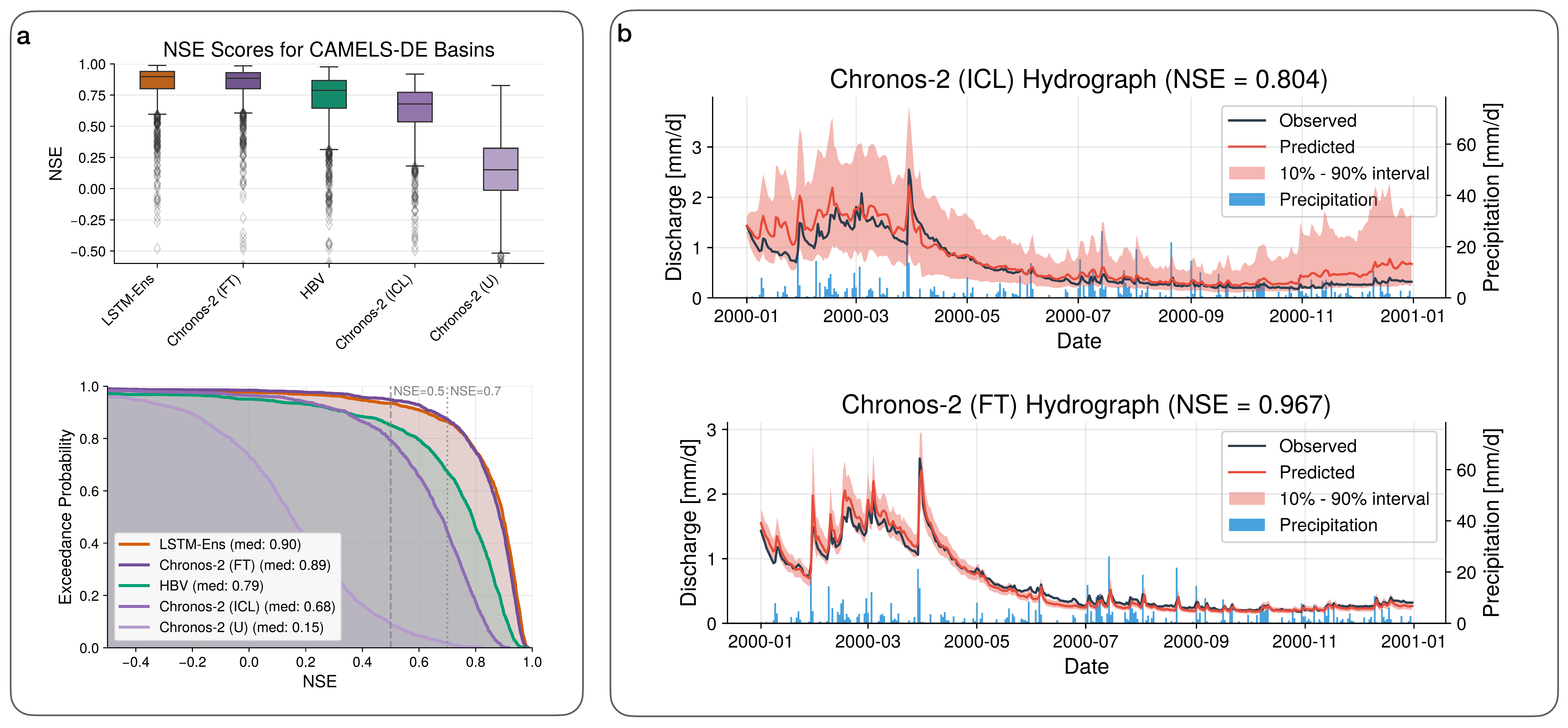}
    \caption{\textbf{Rainfall-runoff simulation using \ourmodel.} Performance is evaluated using the Nash–Sutcliffe Efficiency (NSE), where a score of 1.0 represents perfect correspondence between simulated and observed discharge. (a) Top: Distribution of NSE scores across 1,582 catchments in the CAMELS-DE dataset. The zero-shot \ourmodel (ICL) model achieves a median NSE of 0.68, significantly outperforming univariate inference (\ourmodel (U), NSE=0.15) by leveraging meteorological covariates. While the domain-specific LSTM ensemble achieves higher performance (NSE=0.90), fine-tuning \ourmodel (FT) bridges this gap, matching state-of-the-art performance without ensembling, access to static attributes, or training on the NSE metric. Box plots show the median (center line), interquartile range (box limits), and 1.5$\times$ interquartile range (whiskers). Bottom: Exceedance probability curves showing the cumulative proportion of basins (y-axis) achieving an NSE score greater than or equal to a specific threshold (x-axis). The fine-tuned model closely tracks the reliability profile of the LSTM ensemble. (b) Hydrograph for a single catchment (DE110180) over one year. The zero-shot prediction (top) captures the relationship between rainfall and runoff but exhibits high uncertainty (shaded region denotes 10-90\% prediction interval). Fine-tuning (bottom) tightens the predictive distribution, accurately modeling the discharge.} 
    \label{fig:hydro}
\end{figure}

Rainfall-runoff simulation is a central problem in hydrology, which is critical for reservoir management, flood forecasting, and climate resilience. It studies how rainfall over a watershed is transformed into surface runoff (streamflow) at the basin outlet~\citep{beven2012rainfall}. 
We evaluated \ourmodel for this problem on the Catchment Attributes and Meteorology for Large-sample Studies (CAMELS-DE) dataset~\citep{loritz2024camels}, a high-quality benchmark providing hydro-meteorological time series for 1,582 catchments across Germany.

To contextualize performance, we compared \ourmodel against two domain-specific standards: the HBV model~\citep{bergstrom1976development}, a conceptual physical model calibrated individually for each catchment, and a state-of-the-art Long Short-Term Memory (LSTM) ensemble~\citep{acuna2024bucket}, which represents the current upper bound in hydrological deep learning.
The LSTM baseline is trained on data from all catchments to predict the discharge using precipitation, radiation, temperature, and static catchment attributes (e.g., soil texture, elevation) as inputs.

\Cref{fig:hydro}(a) presents the distribution of Nash–Sutcliffe Efficiency (NSE) scores~\citep{nash1970river}, a metric commonly used to evaluate rainfall–runoff models.
Remarkably, \ourmodel (with ICL) achieves a median NSE of 0.68.
Without ever seeing a hydrograph during training, the model successfully identifies the relationship between precipitation events and streamflow response.
This performance is starkly superior to univariate inference (median NSE = 0.15), confirming that the model is actively leveraging meteorological covariates rather than simply autoregressing on historical discharge.

While zero-shot performance lags behind the highly specialized baselines, \ourmodel provides viable hydrological forecasts instantly, without the training or calibration these baselines require. Lightweight fine-tuning on CAMELS-DE data effectively bridges this gap, with \ourmodel matching the performance of the state-of-the-art LSTM ensemble --- despite using a single model without static catchment attributes or the NSE objective function. \Cref{fig:hydro}(b) illustrates this progression: in zero-shot mode, the model captures the general shape of discharge but exhibits high uncertainty; after fine-tuning, the predictive distribution tightens around the ground truth, effectively learning the rainfall-runoff relationship. This demonstrates \ourmodel's flexibility as both an immediately useful generalist and a robust initialization for specialist models.

\subsection{Hierarchical forecasting}

\begin{figure}[htb]
\centering
\includegraphics[width=\textwidth]{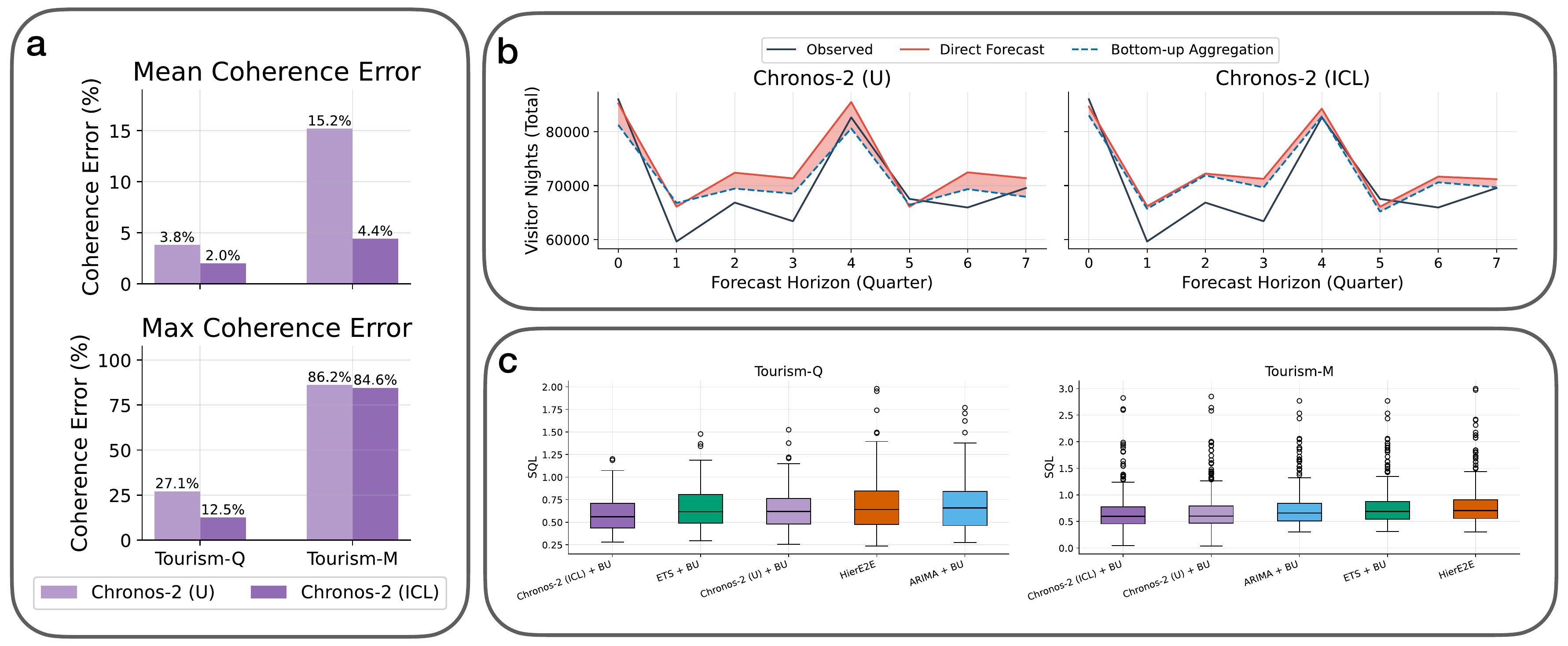}
\caption{\textbf{Hierarchical forecasting with \ourmodel.} (a) Mean and maximum coherence error on the Tourism-Q and Tourism-M datasets for \ourmodel~(U) and \ourmodel~(ICL). Enabling ICL reduces the coherence error on both datasets. (b) Root-level forecasts on Tourism-Q. Solid red line shows the direct forecast generated by the respective models for the root level, while the dashed blue line shows the bottom-up aggregated forecasts. The shaded region shows the coherence gap. \ourmodel~(ICL) produces a narrower gap than \ourmodel~(U) indicating that group attention implicitly learns the aggregation structure. (c) Distribution of per-series scaled quantile loss (SQL) with bottom-up reconciliation on Tourism-Q (left) and Tourism-M (right). \ourmodel~(ICL) achieves the lowest median SQL on both datasets (0.56 and 0.60, respectively), outperforming the end-to-end HierE2E model trained specifically for hierarchical coherence as well as ETS and ARIMA statistical baselines. Methods are ordered by median SQL.}
\label{fig:hierarchical}
\end{figure}

Many real-world forecasting problems exhibit hierarchical structure: forecasts at different levels of aggregation (e.g., national, regional, local) should ideally be mutually consistent~\citep{hyndman2018forecasting}. Producing coherent hierarchical forecasts --- where aggregate predictions equal the sum of their disaggregated components --- typically requires specialized end-to-end models~\citep{rangapuram2021end} or post-hoc reconciliation~\citep{hyndman2011optimal,wickramanayake2019optimal,bentaieb2019regularized}. We investigated whether \ourmodel can address this setting without explicit hierarchy information.

We evaluated on the Australian Tourism benchmarks~\citep{athanasopoulos2009hierarchical,wickramanayake2019optimal}: Tourism-Q (89 series, 4 levels, quarterly) and Tourism-M (555 series, 8 levels, monthly, with a crossed geography$\times$purpose structure). We compared two configurations: \ourmodel~(U), where each series is forecast independently; and \ourmodel~(ICL), where all series in the hierarchy are passed jointly as a multivariate input, enabling group attention to capture cross-level dependencies.

\Cref{fig:hierarchical}(a) shows that \ourmodel~(ICL) produces substantially more coherent forecasts than \ourmodel~(U) without any reconciliation. We measure coherence error as the relative difference between a direct forecast and its bottom-up aggregation from leaf-level forecasts (see Appendix~\ref{sec:methods} for formal definition). Mean coherence error drops from 3.8\% to 2.0\% on Tourism-Q and from 15.2\% to 4.4\% on Tourism-M. \Cref{fig:hierarchical}(b) illustrates this at the root level, where the gap between the direct forecast and the bottom-up aggregation is visibly narrower under ICL. This suggests that group attention implicitly captures aspects of the hierarchical aggregation structure.

To produce perfectly coherent probabilistic forecasts, we applied bottom-up reconciliation (see Appendix~\ref{sec:methods} for details). We chose this approach for simplicity and to isolate the model's intrinsic capabilities; more sophisticated reconciliation methods could also be applied. We compared against HierE2E~\citep{rangapuram2021end}, an end-to-end hierarchical deep learning model trained from scratch on each dataset, and statistical methods (ETS and ARIMA) with the same bottom-up procedure. \ourmodel~(ICL) achieves the best SQL on both datasets (\Cref{fig:hierarchical}c), outperforming competing methods on these benchmarks without any task-specific training or knowledge of the hierarchy.
These results demonstrate that \ourmodel{}'s group attention mechanism captures cross-level dependencies well enough to produce competitive coherent forecasts without task-specific training, explicit hierarchy information, or specialized reconciliation.

\section{Conclusion}\label{sec13}

In this work, we introduced \ourmodel, a time series foundation model designed to handle a wide range of forecasting scenarios --- both univariate and multivariate --- in a zero-shot manner. 
\ourmodel marks a shift from fitting parameters to inferring dynamics.
Unlike traditional approaches that require training bespoke models for every dataset, \ourmodel implicitly infers coupled dynamics purely from historical context.
By training extensively on multivariate synthetic data, we demonstrated that the fundamental grammar of temporal variation can be learned and transferred to real-world complexity, effectively bridging the gap between domains ranging from hydrology to macroeconomics.
\ourmodel democratizes access to advanced forecasting by offering a single pretrained model as an alternative to task-specific development, providing a versatile tool for analyzing coupled dynamics across diverse temporal systems.

Building on this framework, future development of \ourmodel can expand along several critical dimensions. First, scaling the diversity of the synthetic pretraining corpus --- e.g., by incorporating outputs from domain-specific simulators --- can further enhance zero-shot generalization for highly complex dynamics. Second, while the current architecture natively handles time-varying covariates, integrating static metadata (e.g., soil composition or demographic profiles) is an important frontier. This capability is vital for extreme \emph{cold-start} scenarios, e.g., the grand challenge of prediction in ungauged basins (PUB) in hydrology, where streamflow must be inferred entirely from meteorological drivers and static basin properties without historical target data. In addition, since critical exogenous signals are often qualitative (e.g., central bank statements or epidemiological reports), extending the model to support multimodal natural language inputs is a promising direction. As these foundation models assume a larger role in science and decision-making, advancing their interpretability is also essential. This includes exploring methods to extract explicit causal graphs or coupling mechanisms directly from the model's internal representations. Lastly, the universal temporal representations learned by \ourmodel hold immense potential for broader downstream applications beyond forecasting, including anomaly detection, classification, and clustering.

\section*{Acknowledgements}
 We thank the developers of open-source libraries used in the development of \ourmodel, including but not limited to \texttt{torch}~\citep{paszke2019pytorch}, \texttt{numpy}~\citep{harris2020array}, \texttt{pandas}~\citep{reback2020pandas,mckinney-proc-scipy-2010}, \texttt{statsmodels}~\citep{seabold2010statsmodels}, \texttt{transformers}~\citep{wolf-etal-2020-transformers}, \texttt{gluonts}~\citep{alexandrov2020gluonts}, \texttt{autogluon}~\citep{shchur2023autogluon}, \texttt{statsforecast}~\citep{garza2022statsforecast}, \texttt{einops}~\citep{rogozhnikov2022einops} and \texttt{scikit-learn}~\citep{pedregosa2011scikit}. We thank Caner Turkmen, Tam Nguyen, Suprabhath Kalahasti, Matteo Chinazzi, Elliot Beck, Mischa Lisovyi, Imry Kissos, Devamanyu Hazarika, Tim Januschowski, Jan Gasthaus, William Gilpin, Annan Yu, Andrew Gordon Wilson and Michael W. Mahoney for their feedback.

\bibliography{main}
\bibliographystyle{tmlr}

\clearpage
\appendix
\section{Methods}\label{sec:methods}

\subsection{Problem statement}

Time series forecasting aims to predict future values of a temporal sequence given historical observations.
Formally, let $\mY_{1:T} = \left[\rvy_1, \dots, \rvy_{T}\right]$ denote a historical time series of length $T$, where each observation $\rvy_t \in \mathbb{R}^D$ can either be univariate ($D=1$) or multivariate ($D>1$).
Given this historical context, the goal is to predict the next $H$ time steps $\mY_{T+1:T+H}$, where $H$ defines the forecast horizon.
Forecasts may be supported by covariates (also known as \emph{exogenous variables}) $\mX_{1:T+H} = \left[\rvx_1, \dots, \rvx_{T+H}\right]$, where $\rvx_t \in \mathbb{R}^M$ represents additional information that can span both historical ($t \leq T$) and future ($t > T$) time steps.
The task itself can be defined as either \emph{point forecasting}, where the objective is to predict a single future value at each time step, or \emph{probabilistic forecasting}, where the objective is to estimate the conditional distribution $\mathcal{P}(\mY_{T+1:T+H} \mid \mY_{1:T}, \mX_{1:T+H})$ in order to capture forecast uncertainty.
\emph{Zero-shot forecasting} refers to the setting in which a model generates forecasts for previously unseen time series datasets without requiring any additional training, adaptation, or fine-tuning.

\subsection{\ourmodel architecture}

\ourmodel is a 120M parameter encoder-only transformer~\citep{vaswani2017attention} model which closely follows the design of the T5 encoder~\citep{raffel2020exploring}, building on prior versions of the Chronos family. \Cref{tab:chronos-comparison} summarizes the key differences between \ourmodel{}, Chronos~\citep{ansari2024chronos}, and Chronos-Bolt~\citep{ansari2024fast}.
In the following, we discuss the key architectural components of the \ourmodel pipeline.

\paragraph{Input construction}
The model operates on two inputs derived from the target $\mY_{1:T}$ and covariates $\mX_{1:T+H}$.
We concatenate all historical values into $\mV = [\rvv_1, \dots, \rvv_T]$, where each $\rvv_t \in \mathbb{R}^{D+M}$ consists of the target observation $\rvy_t$ and the corresponding covariate vector $\rvx_t$.
Similarly, we define the future values as $\mW = [\rvw_{T+1}, \dots, \rvw_{T+H}]$, where $\rvw_t \in \mathbb{R}^{D + M}$ contains known (future) covariate values $\rvx_{t}$ when available, while the entries corresponding to targets and past-only covariates are set to missing values.

Categorical covariates in $\mX_{1:T+H}$ are transformed into real-valued representations before being concatenated into $\mV$ and $\mW$. 
For univariate targets, we apply target encoding \citep{pedregosa2011scikit,micci2001preprocessing}, which maps each category to a numerical value based on its relationship with the target. 
For multivariate targets, the model falls back to ordinal encoding, assigning a unique integer to each category.

\paragraph{Robust scaling} The input values, $\mV$ and $\mW$, may be at an arbitrary scale, so our tokenization pipeline begins by normalizing the series.
We adopt \emph{standardization}, a widely used normalization method in the literature, and introduce an additional step: applying the $\sinh^{-1}$ transformation to the standardized values.
This log-like transformation further stabilizes variance and reduces the influence of outliers on the objective function.
Formally, each historical value $v_{t,d}$ and the future value $w_{t,d}$ are normalized as
\begin{align}
\tilde{v}_{t,d} &= \sinh^{-1}\!\left(\frac{v_{t,d} - \mu_d}{\sigma_d}\right) \text{for } t \in \{1,\dots,T\}\text{ and } d \in \{1,\dots,D+M\}, \label{eq:normalization}\\
\tilde{w}_{t,d} &= \sinh^{-1}\!\left(\frac{w_{t,d} - \mu_d}{\sigma_d}\right) \text{for } t \in \{T+1,\dots,T+H\}\text{ and } d \in \{1,\dots,D+M\},
\end{align}

\noindent where $\mu_d$ and $\sigma_d$ are the mean and standard deviation of the historical values $[v_{1,d}, ..., v_{T,d}]$, respectively.
Any missing values in $\mV$ are excluded when computing $\mu_d$ and $\sigma_d$.
The normalized historical values $\tilde{\mV}$ and future values $\tilde{\mW}$ are concatenated to construct the input matrix $\mU = [\tilde{\mV}, \tilde{\mW}] \in \R^{(T+H) \times (D +M)}$.

\paragraph{Meta features} 
During tokenization, each dimension of $\mU$ is processed independently by the model. 
To describe the tokenization procedure, consider a single column $\vu_d = [u_{1,d}, \dots, u_{T+H,d}]^\top$ corresponding to one target or covariate dimension $d$.
Two additional meta features are appended to each column: a time index and a mask. The time index
$
    \rvj = \left[-\frac{T}{C}, -\frac{T-1}{C}, \dots, 0, \dots, \frac{H-1}{C}\right]
$
encodes the relative position of each time step,
where $C$ is the maximum context length supported by the model.
It provides explicit information about temporal ordering to the model which is beneficial when using patch-based inputs.
The mask $\rvm_d$ is a binary indicator equal to 1 when the value is observed, and 0 otherwise.
It serves two purposes: indicating which values are missing in the historical context and specifying which input dimensions correspond to known (future) covariates.
After construction of the mask, all missing values in $\vu_d$ are replaced with zeros.

\paragraph{Patching and embedding} The input $\rvu_d$ with the corresponding meta features, $\rvj$ and $\rvm_d$, are split into non-overlapping patches of length $P=16$~\citep{Nie2023PatchTST}.
The context and future sections of the time series and meta features are split into patches separately.
When $T$ and $H$ are not multiples of $P$, zero padding is applied on the left (context) or right (future).
Let $\overline{\rvu}_p$, $\overline{\rvj}_p$, and $\overline{\rvm}_p$ denote the $p$-th patches of the input, time index, and mask, respectively.
These are concatenated and mapped into the embedding space using a residual network, $f^{\mathrm{in}}_\phi: \R^{3P} \to \R^{\dmodel}$,
\begin{equation}
    \rvh_{p} = f^{\mathrm{in}}_\phi\left(\left[\overline{\rvu}_p, \overline{\rvj}_p, \overline{\rvm}_p\right]\right),
\end{equation}
where $\phi$ denotes parameters of the residual network and $\dmodel$ is the hidden dimension of the transformer model.
Between the patch embeddings of the context and future, we include a special \texttt{REG} token which serves both as a separator token and an attention sink~\citep{xiao2023efficient}. 

\paragraph{Time attention} The time attention layer is the usual attention layer found in typical sequence models. It applies self-attention along the temporal axis and aggregates information across patches of the same input dimension. We replace relative position embeddings used in the self-attention layers of the original T5 model with rotary position embeddings (RoPE)~\citep{su2024roformer} which have become the de-facto standard for position embeddings in modern transformer-based models~\citep{touvron2023llama}.

\paragraph{Group attention}
The \emph{group attention} layer is central to enabling the in-context learning capabilities of \ourmodel. This layer aggregates information across time series that belong to the same group at a given patch index. A group refers to a set of related time series and may refer to different things depending on the forecasting task. 

Within a batch of size $B$, multiple groups of varying sizes are possible, each identified by group IDs $\vg$, a vector of length $B$.
Internally, the group attention layer maps these IDs to a two-dimensional attention mask, ensuring that aggregation occurs only within groups and not across them. Since time series within a group lack a natural ordering, the group attention layer omits positional embeddings.

\paragraph{Quantile head}

After a sequence of alternating time and group attention layers, the embeddings of future patches of the $D$ target dimensions are passed through a residual block to produce the direct multi-step quantile forecast $\hat{\mZ} \in \mathbb{R}^{H \times D \times |\gQ|}$.
By producing forecasts for multiple target patches within a single forward pass, the model can efficiently generate predictions over long forecast horizons.
\ourmodel predicts a set of 21 quantiles $\gQ = \{0.01, 0.05, 0.1, \dots, 0.9, 0.95, 0.99\}$. 
Any queried quantiles falling outside this set are obtained through linear interpolation.

\subsection{Training}

During training, batches are constructed to include heterogeneous forecasting tasks: univariate forecasting, multivariate forecasting (which also covers tasks with past-only covariates), and multivariate forecasting with known (future) covariates. Each task is characterized by the number of target dimensions $D$, the number of covariates $M$, and the role of each dimension (target, past-only covariate, or known covariate).
A unique group ID is assigned to each task, and the combination of group IDs $\vg$ with whether the future input $\mW$ is observed allows the model to infer the specific forecasting setup.

The model is trained using the quantile regression objective
\begin{equation}
   \sum_{q \in \gQ} \Big(q \cdot \max(z - \hat{z}^q, 0) + (1 - q) \cdot \max(\hat{z}^q - z, 0)\Big),
\end{equation}
where $\hat{z}^q$ is the forecast at quantile level $q$, and $z$ is the corresponding target value normalized as in~\eqref{eq:normalization}. 
The loss is averaged over all forecast steps and items in the batch and is computed only on target dimensions, with entries corresponding to known (future) covariates or missing target values excluded from the objective.
The number of output patches is randomly sampled for each batch during training. 

The model is trained in two stages using the Adam optimizer with weight decay of 0.1 and maximum gradient norm set to 1. First, the model is pretrained with a maximum context length of 2048, 12 maximum output patches and learning rate of $10^{-4}$ for 2 million gradient update steps. In the second stage, the context length is extended to 8192, the maximum number of sampled output patches is increased to 64 and the learning rate is reduced to $10^{-5}$, and the model is trained for an additional 100 thousand steps. In both stages, we train on $4\times$A100 GPUs (40GB VRAM) with a batch size of 128 time series per GPU, resulting in an effective batch size ($B$) of 512 time series. The first stage uses distributed data parallel (DDP), while the second stage uses fully-sharded data parallel (FSDP) with activation checkpointing to accommodate longer contexts. The model acquires its core capabilities in the first stage; the second stage enables extrapolation to longer contexts and prediction horizons. Longer contexts enable the model to capture long-term seasonalities in high-frequency time series, while multi-patch outputs allow for long-horizon forecasts without relying on heuristics.

\subsection{Training data}

The training corpus combines real and synthetic univariate datasets (Supplementary Table~\ref{tab:real-uni-datasets}) with synthetic multivariate data as discussed in Section~\ref{sec:data}.

Our multivariate synthetic data generation process begins with univariate generators capable of producing a rich diversity of temporal patterns. A randomly chosen number of time series are first sampled from these univariate generators. A function is then applied to these univariate series to introduce correlations between them.

The set of univariate generators includes: (i) \textbf{KernelSynth}~\citep{ansari2024chronos}, which generates time series using Gaussian processes (GP) by randomly combining basic kernels (such as linear and periodic) through addition and multiplication to construct complex composite kernels; (ii) \textbf{Trend-Seasonality-Irregularity (TSI)}~\citep{bahrpeyma2021methodology}, which exploits typical time series decomposition to construct series by sampling trends, seasonalities, and irregularities from a suite of predefined components and combining them through an additive or multiplicative model; (iii) \textbf{AutoRegression (AR)}~\citep{box2015time}, which constructs a random order-$p$ autoregressive model and simulates it to generate time series; and (iv) \textbf{Exponential Smoothing (ETS)}~\citep{hyndman2008forecasting}, which simulates time series from a randomly constructed exponential smoothing model.

With time series sampled from the aforementioned univariate generators, we construct multivariate data by applying suitable random transformation functions which introduce instantaneous or time-delayed relationships between the time series.

Instantaneous transformations are applied across the univariate series at the same time-step. We use two types of instantaneous transformation functions: linear and non-linear. The linear function is realized by a simple transformation matrix which may be dense, sparse, or a sign matrix, or may encode other relationships such as aggregation, where the number of output time series is larger than the number of input series and the additional outputs denote summations of the input series. The random non-linear transformation is characterized by a multi-layer perceptron (MLP), where the number of layers, type of nonlinearities, and number of hidden units are randomly sampled.

Time-delayed transformations are functions which take a window of the last $k$ time-steps from the univariate time series as inputs and are recursively applied to yield multivariate time series with time-varying relationships. As before, we use linear and non-linear transformation functions. The linear function is characterized by a temporal causal model (TCM). The algorithm first selects a subset of the univariate series as exogenous variables --- they mimic external factors which affect the system but are not affected by it. A set of edges across time and variables are then sampled to denote relationships, with edge weights sampled from a truncated normal distribution. This yields the parameters of a vector autoregressive (VAR) process~\citep{lutkepohl2013introduction} which is then applied to the univariate series. For non-linear time-delayed relationships, we apply a recurrent neural network (RNN) to the univariate time series, where the number and type of layers are randomly sampled.

In total, the training corpus comprises approximately 50 million real univariate time series (augmented via TSMixup~\citep{ansari2024chronos} using the original time series in \Cref{tab:real-uni-datasets}) with lengths ranging from 64 to 4096, and 10 million synthetically generated univariate series of length 4096. The multivariate synthetic corpus consists of approximately 7 million sequential transformation series and 5 million instantaneous transformation series, each of length 2048 with 2 to 32 variates. For long-context training (Stage 2), we additionally generate 10 million TSI~\citep{bahrpeyma2021methodology} series of length 12,288. During training, univariate and multivariate tasks are sampled in equal proportion, with context lengths sampled randomly up to the maximum for each stage.

\subsection{Inference}
\label{sec:inference}

Forecasts are generated by de-normalizing the model predictions $\hat{z}_{t,d}^q$ and applying the inverse transformation of \eqref{eq:normalization}.
Formally, the quantile head output $\hat{z}_{t,d}^q$ is transformed as
\begin{align}
    \hat{y}_{t,d}^q &= \mu_d + \sigma_d \cdot \sinh({\hat{z}_{t,d}^q}),
\end{align}
to obtain the prediction $\hat{y}_{t,d}^q$ of the quantile level $q$ at time step $t$ along the target dimension $d$.

During inference, multiple time series in a batch can be grouped to solve different forecasting tasks:
\begin{itemize}
    \item \emph{univariate forecasting}: each item in the batch is assigned a unique group ID. This ensures that the model makes independent predictions for each time series in the batch.
    \item \emph{multivariate forecasting}: each variate which belongs to the same multivariate series is assigned the same group ID with variates from different multivariate series having distinct group IDs. This allows the model to share dynamics information between different variates of a multivariate time series.
    \item \emph{forecasting with covariates}: all target(s), past-only and known covariates belonging to the same task are assigned the same group ID. The future inputs $\mW$ corresponding to known covariates contain their known future values. The predictions generated by the model for covariates are ignored.
\end{itemize}

\Cref{tab:forecasting-tasks} summarizes how group IDs and future inputs must be specified to solve different forecasting tasks. In addition to these, \ourmodel can also be used in the \emph{full cross learning} mode where each item in the batch is assigned the same group ID regardless of whether the item is a target, a past-only covariate or a known covariate.
Since each item belongs to the same group, the model shares information across items in the batch and makes joint predictions for the entire batch.

\subsection{Evaluation protocol for standard benchmarks}
In this section, we detail our evaluation protocol including the benchmark datasets, baselines and metrics used in the experiments.

\paragraph{Benchmarks}

We evaluated \ourmodel on three established forecasting benchmarks: \fevbench~\citep{shchur2025fev}, \chronosbenchii~\citep{ansari2024chronos}, and \gifteval~\citep{aksu2024gifteval}. We strictly adhered to the data splits and evaluation protocols defined in the respective publications.
\fevbench provides comprehensive coverage of diverse real-world scenarios, comprising 100 forecasting tasks (68 of which involve multivariate dependencies).
\chronosbenchii, originally proposed to evaluate early Chronos models~\citep{ansari2024chronos}, consists of 27 tasks characterized predominantly by short historical contexts (averaging fewer than 300 time steps).
\gifteval focuses on high-frequency time series and long-horizon forecasting, containing 97 tasks derived from 55 distinct datasets.

\paragraph{Baselines}

We compared \ourmodel against a range of leading task-specific models (trained individually on each benchmark task) and pretrained foundation models.
Task-specific baselines included statistical methods (AutoETS, AutoARIMA, AutoTheta, and their ensemble~\citep{hyndman2018forecasting, petropoulos2020simple}), tree-based methods (LightGBM~\citep{ke2017lightgbm}, CatBoost~\citep{prokhorenkova2018catboost}), and deep learning architectures (DeepAR~\citep{salinas2020deepar}, TFT~\citep{lim2021temporal}, PatchTST~\citep{Nie2023PatchTST}).
Foundation model baselines included TiRex~\citep{auer2026tirex}, TimesFM-2.5~\citep{das2024decoder}, Toto-1.0~\citep{cohen2025time}, Chronos-Bolt~\citep{ansari2024chronos}, TabPFN-TS~\citep{hollmann2025tabpfn, hoo2025tables}, Moirai-1.1~\citep{woo2024unified}, Moirai-2.0~\citep{liu2025moirai}, Granite-TTM-R2~\citep{ekambaram2024tinytimemixersttms}, and TimePFN~\citep{taga2025timepfn}.

Where available, we utilized reported baseline results from the original benchmarking papers; otherwise, we reproduced results using standard libraries.
Statistical models were implemented via StatsForecast~\citep{garza2022statsforecast} (max. runtime 12 hours/task) on AWS \texttt{m6i.4xlarge} instances (16 vCPUs, 64\,GB RAM).
Tree-based models were implemented via MLForecast; to address sensitivity to preprocessing, we tuned differencing, scaling, and lag transformations for each task, selecting lag/time features via frequency-based heuristics.
Deep learning models were implemented via AutoGluon~\citep{shchur2023autogluon} on AWS \texttt{g5.2xlarge} instances (A10G GPU, 24\,GB VRAM). To prevent under-fitting, we extended the training budget to 1,000 epochs (with early stopping) and a 24-hour maximum runtime.
Task-specific models were re-trained for each evaluation window to account for distribution shifts. Foundation models were evaluated using their respective official libraries on \texttt{g5.2xlarge} instances.

\paragraph{Metrics and aggregation strategy}

\textbf{Point and probabilistic forecast accuracy}. Point forecast accuracy was evaluated using the Mean Absolute Scaled Error (MASE), which normalizes the mean absolute error by the historical seasonal error. For a task with $N$ time series, each with $D$ dimensions and forecast horizon $H$, MASE is defined as:
\begin{equation}
    \operatorname{MASE} = \frac{1}{N} \frac{1}{D} \frac{1}{H} \sum_{n=1}^{N} \sum_{d=1}^{D} \frac{1}{a_{n,d}} \sum_{t=T+1}^{T+H} |y_{n,t,d} - \hat{y}_{n,t,d}|,
\end{equation}
where $y_{n,t,d}$ is the ground truth value, $\hat{y}_{n,t,d}$ is the point forecast, and $a_{n,d}$ is the historical absolute seasonal error of series $n$ along dimension $d$:
\begin{equation}
    a_{n,d} = \frac{1}{T-m} \sum_{t=m+1}^T |y_{n,d,t} - y_{n,d,t-m}|,
\end{equation}
and $m$ is the seasonal period.

Probabilistic forecast quality was measured using the Scaled Quantile Loss (SQL), which aggregates the pinball loss across quantile levels, normalized by the same seasonal error $a_{n,d}$:
\begin{equation}
  \operatorname{SQL} = \frac{1}{N} \frac{1}{D} \frac{1}{H} \sum_{n=1}^{N} \frac{1}{a_{n,d}} \sum_{t=T+1}^{T+H} \sum_{q}  \rho_q(y_{n,d,t}, \hat{y}^q_{n,d,t}),
\end{equation}
where $\hat{y}_{n,d,t}^q$ denotes the predicted quantile $q$ for dimension $d$ of time series $n$ at time $t$, and the quantile loss $\rho_q$ is defined as:
\begin{equation}
    \rho_q(y, \hat{y}^q) = \begin{cases}
      2 \cdot (1 - q) \cdot (\hat{y}^q - y), & \text{if } y < \hat{y}^q\\
      2 \cdot q \cdot (y - \hat{y}^q ), & \text{if } y \ge \hat{y}^q.
      \end{cases}
\end{equation}
Consistent with the original benchmarking protocols, we used the Weighted Quantile Loss (WQL) for \chronosbenchii and \gifteval. Both SQL and WQL approximate the Continuous Ranked Probability Score (CRPS); however, unlike SQL, WQL depends on the scale of the time series.

\textbf{Aggregating benchmark results.} After evaluating $M$ models on $R$ tasks, we obtain the error matrix $E \in \mathbb{R}_{\ge 0}^{R \times M}$, where $E_{rj}$ denotes the error of model $j$ on task $r$. Lower values correspond to more accurate forecasts. We aggregate these results following the methodology in \citet{shchur2025fev}.
The \emph{average win rate} $W_j$ measures the probability that model $j$ outperforms another randomly selected model on a randomly selected task:
\begin{equation}
    W_j =
    \frac{1}{R(M-1)}
    \sum_{r=1}^{R}
    \sum_{k \neq j}
    \left(
        \mathbb{I}[E_{rj}<E_{rk}]
        + \frac{1}{2}\mathbb{I}[E_{rj}=E_{rk}]
    \right),
\end{equation}
where $\mathbb{I}[\cdot]$ is the indicator function.
The \emph{skill score} $S_j$ measures the average relative improvement of model $j$ over the Seasonal Naive baseline $\beta$:
\begin{equation}
    S_j =
    1 -
    \exp\left(
        \frac{1}{R}
        \sum_{r=1}^{R}
        \log\left(
            \max\left(
                10^{-2},
                \min\left(E_{rj}/E_{r\beta},\,100\right)
            \right)
        \right)
    \right).
\end{equation}
The average win rate and skill score are mathematically equivalent to the average rank and geometric mean relative error used in prior work~\citep{ansari2024chronos,aksu2024gifteval}, but are more directly interpretable. Statistical significance of \ourmodel's performance against each baseline was assessed using the Wilcoxon signed-rank test.

Additionally, we report the \emph{improvability} metric $I_j$ introduced in \citet{erickson2026tabarena}, which measures the average percentage by which the error of model $j$ can be reduced by selecting the best-in-hindsight model on each task:
\begin{equation}
    I_j =
    \frac{100}{R}
    \sum_{r=1}^{R}
    \frac{E_{rj}-\min_k E_{rk}}{E_{rj}}.
\end{equation}
The improvability ranges from $0\%$ to $100\%$, where lower values indicate that a model is consistently closer to the per-task best, even when it is not the top performer on every individual task.

\textbf{Calibration evaluation.}
To assess the calibration of probabilistic forecasts, we report the Weighted Interval Score (WIS)~\citep{gneiting2007strictly} and the Mean Absolute Coverage Error (MACE). Both metrics are computed on three central prediction intervals: 90\% $[0.05,0.95]$, 95\% $[0.025,0.975]$, and 98\% $[0.01,0.99]$.
The WIS is a strictly proper scoring rule for evaluating prediction intervals that jointly assesses calibration and sharpness~\citep{gneiting2007strictly}. We report the geometric mean of the WIS relative to the Seasonal Naive baseline across tasks, where lower values indicate better performance.
The MACE measures the average absolute discrepancy between nominal and empirical interval coverage. Unlike the WIS, it is not a proper scoring rule and evaluates calibration alone, but provides an intuitive measure of how closely the empirical coverage matches the nominal coverage. Lower values indicate better calibration.

\subsection{Ablation studies}

In this section, we present experiments that disentangle the impact of different design choices, including training data composition, group attention, the two-stage training protocol for long-context capabilities, and other training hyperparameters.

\paragraph{Training data composition}

We trained \ourmodel on different data compositions using the single-stage training protocol and compared against the first-stage checkpoint of the reference \ourmodel. \Cref{tab:ablations} shows that training only on real data leads to significant degradation in forecast accuracy across all three benchmarks. Since this configuration effectively trains the model on univariate data only --- all multivariate data used to train \ourmodel is synthetically generated --- we observe a particularly large drop on \fevbench, which focuses on multivariate tasks including those with covariates.

The model trained only on synthetic data delivers strong performance across all three benchmarks, in some cases only marginally worse than the reference model. These results underscore the importance of synthetic data and validate the design choice of relying on synthetic generation for multivariate pretraining rather than real multivariate datasets, which are scarce and would overlap with evaluation benchmarks. We further investigated the impact of removing specific types of synthetic multivariate data. Both models with sequential or instantaneous multivariate data removed perform worse than the reference model but better than the model trained on real data only. Furthermore, removing instantaneous multivariate data yields worse results on \fevbench than removing sequential data. This is consistent with the composition of \fevbench, which contains a substantial proportion of multivariate tasks with known future covariates where interactions between time series are typically instantaneous.

\paragraph{Group attention}

Group attention is the key architectural component that enables the multivariate and cross-learning capabilities of \ourmodel. \Cref{fig:main-results} shows that disabling group attention at inference time dramatically worsens forecast accuracy across the three benchmarks, confirming its critical role in leveraging multivariate relationships during prediction. Additionally, we trained an ablated variant where group attention is disabled during training by using an identity attention mask in the group attention layers, ensuring that no information flows between different time series. \Cref{tab:ablations} shows that this model underperforms the reference, especially on \fevbench where group attention enables more accurate forecasts on multivariate tasks. Together, the inference-time (\Cref{fig:main-results}) and training-time (\Cref{tab:ablations}) ablations confirm that group attention is essential both for learning multivariate representations and for exploiting them at inference time.

\paragraph{Long-context training}

As described in the training protocol, \ourmodel is trained in two stages: first with a context length of 2048 and subsequently with 8192. \Cref{tab:ablations} shows the improved accuracy of the final model after long-context training. The improvements are primarily attributable to higher-frequency tasks (e.g., 5-minute and 1-minute resolution), where longer context captures additional seasonal structure useful for accurate predictions.

\paragraph{Learning rate}

In Stage-1 training of \ourmodel, we use a peak learning rate of $1\times10^{-4}$ and vary this reference value in \Cref{tab:ablations}.
Across the tested range ($1\times10^{-5}$ to $1\times10^{-3}$), forecast accuracy varies to a limited extent on all three benchmarks.
The reference learning rate offers the best overall trade-off.

\paragraph{Patch size}
Using the Stage-1 training protocol, we ablate the effect of the input and output patch size, keeping the stride equal to the patch size.
\Cref{tab:ablations} shows that increasing the patch size to 32 and 64 degrades accuracy consistently across all three benchmarks.
Decreasing the patch size to 8 is competitive with the reference on \fevbench and \chronosbenchii but clearly worse on \gifteval, indicating that smaller patches do not uniformly help.
Overall, the reference patch size of 16 offers the best trade-off across benchmarks.

\begin{table}
    \centering
    \resizebox{\textwidth}{!}{%
    \begin{tabular}{lcccccc}
    \toprule
     & \multicolumn{2}{c}{\fevbench} & \multicolumn{2}{c}{\chronosbenchii} & \multicolumn{2}{c}{\gifteval} \\
    \cmidrule(lr){2-3} \cmidrule(lr){4-5} \cmidrule(lr){6-7}
    Model & SQL $\uparrow$ & MASE $\uparrow$ & WQL $\uparrow$ & MASE $\uparrow$ & WQL $\uparrow$ & MASE $\uparrow$ \\
    \midrule
    \multicolumn{7}{l}{\textit{Training data composition}} \\
    \addlinespace[2pt]
    \ourmodel \textit{(Stage-1)} & \textbf{0.469} & \textbf{0.352} & \textbf{0.458} & \textbf{0.250} & \textbf{0.501} & \textbf{0.281} \\
    \addlinespace[2pt]
    \quad Real data only & 0.419 & 0.289 & 0.431 & 0.233 & 0.484 & 0.258 \\
    \quad Synthetic data only & 0.449 & 0.329 & 0.453 & 0.246 & 0.491 & 0.259 \\
    \quad No sequential multivariate data & 0.459 & 0.339 & 0.447 & 0.232 & 0.493 & 0.268 \\
    \quad No instantaneous multivariate data & 0.449 & 0.328 & 0.453 & 0.254 & 0.486 & 0.271 \\
    \midrule
    \multicolumn{7}{l}{\textit{Group attention}} \\
    \addlinespace[2pt]
    \ourmodel \textit{(Stage-1)} & \textbf{0.469} & \textbf{0.352} & \textbf{0.458} & \textbf{0.250} & \textbf{0.501} & \textbf{0.281} \\
    \addlinespace[2pt]
    \quad No group attention & 0.436 & 0.311 & 0.441 & 0.234 & 0.490 & 0.272 \\
    \midrule
    \multicolumn{7}{l}{\textit{Long context training}} \\
    \addlinespace[2pt]
    \ourmodel &\textbf{ 0.473} & \textbf{0.355} & \textbf{0.466} & \textbf{0.265} & \textbf{0.514} & \textbf{0.302} \\
    \addlinespace[2pt]
    \quad Without long context training\textsuperscript{\textdagger} & 0.469 & 0.352 & 0.458 & 0.250 & 0.501 & 0.281 \\
    \midrule
    \multicolumn{7}{l}{\textit{Learning rate}} \\
    \addlinespace[2pt]
    \ourmodel \textit{(Stage-1)}, $1\times10^{-4}$ & \textbf{0.469} & \textbf{0.352} & \textbf{0.458} & \textbf{0.250} & \textbf{0.501} & \textbf{0.281} \\
    \addlinespace[2pt]
    \quad $1\times10^{-5}$ & 0.455 & 0.334 & 0.463 & 0.265 & 0.491 & 0.259 \\
    \quad $3\times10^{-5}$ & 0.464 & 0.347 & 0.455 & 0.247 & 0.494 & 0.274 \\
    \quad $3\times10^{-4}$ & 0.461 & 0.345 & 0.460 & 0.255 & 0.496 & 0.280 \\
    \quad $1\times10^{-3}$ & 0.455 & 0.335 & 0.448 & 0.252 & 0.492 & 0.269 \\
    \midrule
    \multicolumn{7}{l}{\textit{Patch size}} \\
    \addlinespace[2pt]
    \ourmodel \textit{(Stage-1)}, 16 & \textbf{0.469} & \textbf{0.352} & \textbf{0.458} & \textbf{0.250} & \textbf{0.501} & \textbf{0.281} \\
    \addlinespace[2pt]
    \quad 8 & 0.461 & 0.350 & 0.474 & 0.282 & 0.428 & 0.247 \\
    \quad 32 & 0.449 & 0.328 & 0.440 & 0.250 & 0.491 & 0.277 \\
    \quad 64 & 0.432 & 0.308 & 0.427 & 0.234 & 0.482 & 0.257 \\
    \bottomrule
    \end{tabular}%
    
    }
    \caption{\textbf{Ablation study on training data composition, group attention, and long-context training.} All metrics are reported as skill scores (higher is better). For each ablation group, the reference model is shown in bold. Training data ablations and the group attention ablation use the first stage (Stage-1) training protocol. \textsuperscript{\textdagger}Equivalent to the Stage-1 checkpoint.}
    \label{tab:ablations}
  \end{table}

\subsection{Details on case studies}
In this section, we provide the details for the various real-world case studies ranging from epidemiology and macro-economics to earth sciences and hydrology, as well as the hierarchical forecasting experiment.

\paragraph{Infectious disease dynamics}
We utilized incidence data for Acute Respiratory Infection (ARI) and Influenza-Like Illness (ILI) spanning October 2024 to January 2026, sourced from the European Centre for Disease Prevention and Control (ECDC) via the RespiCast repository~\citep{ecdc_respicast}.
To preclude look-ahead bias from data revisions, \ourmodel was evaluated on historical data snapshots, strictly adhering to the RespiCast protocol.
We evaluated two configurations: \ourmodel (U), a univariate mode where each country's incidence is forecast in isolation; and \ourmodel (ICL), an in-context learning mode where forecasts are generated jointly across countries.
Baseline forecasts for the quantile model and the RespiCast Hub-Ensemble were obtained directly from the RespiCast snapshots.

\paragraph{Consumer price index forecasting}
Data for German Consumer Price Index (CPI) components were retrieved from GENESIS-Online, the central database of the German Federal Statistical Office, on January 19, 2026.
We selected 110 components at Level 4 of the Classification of Individual Consumption According to Purpose (COICOP) for evaluation.
A rolling window evaluation was conducted annually from 2021 to 2025, capturing the post-COVID-19 economic recovery period.
Baselines included AutoARIMA, AutoETS, and DynOptTheta (implemented via StatsForecast), as well as a Bayesian VAR (BVAR) with Minnesota-style shrinkage~\citep{giannone2015prior}, estimated separately for each COICOP division on monthly log-changes with hyperparameters selected via marginal likelihood maximization.
For \ourmodel (U), items were forecast independently; for \ourmodel (ICL), items within the same Level 2 COICOP group were predicted jointly.

\paragraph{Carbon flux prediction}

We utilized daily eddy-covariance measurements of carbon and energy fluxes (NEE, LE, H) alongside meteorological drivers (TA, SW\_IN, VPD) from the FLUXNET dataset~\citep{pastorello2020fluxnet2015}.
Analysis was restricted to sites with high-quality ($>$80\% quality control flag), continuous observation coverage across the context and evaluation periods.
Evaluations included a static summer window (June--August 2020) and a monthly rolling forecast for the entire year 2020, utilizing a 12-month historical context to predict the subsequent month.
We evaluated two zero-shot configurations: \ourmodel (U), utilizing only the isolated target flux history; and \ourmodel (ICL), conditioned jointly on the historical fluxes and future meteorological covariates.
Supervised baselines consisted of pooled and site-specific XGBoost regressors~\citep{chen2016xgboost} trained on available historical data.
Forecast accuracy was evaluated using RMSE and normalized RMSE.

\paragraph{Rainfall-runoff simulation}
We utilized the CAMELS-DE dataset for hydrological modeling.
Pre-computed simulations for the HBV and LSTM-Ensemble baselines were obtained directly from the dataset.
These baselines were calibrated on hydrometeorological data spanning 1970 to 1999.
For this study, we selected the subsequent year (2000) as the evaluation period.
\ourmodel was conditioned on historical context from 1970 to 1999 to predict streamflow for the year 2000, utilizing precipitation, radiation, and temperature as future covariates.
To ensure fair comparison, \ourmodel was fine-tuned on the same 1970--1999 data interval used for the baselines. Fine-tuning was performed using the Adam optimizer with a learning rate of $10^{-4}$ for 10,000 gradient update steps.

\paragraph{Hierarchical forecasting}

We utilized the Australian domestic tourism datasets~\citep{athanasopoulos2009hierarchical,wickramanayake2019optimal}, standard benchmarks for hierarchical forecasting. Tourism-Q is a quarterly hierarchical dataset with 89 series organized in 4 levels (Total $\to$ Purpose $\to$ State $\to$ City/Non-city), spanning 1998--2006 (36 observations per series, $H{=}8$). Tourism-M is a monthly grouped dataset with 555 series in a crossed geography$\times$purpose structure (8 levels, 304 bottom-level series), spanning 1998--2016 (228 observations per series, $H{=}12$).

The coherence error metric follows \citet{rangapuram2021end}: for each aggregate series $i$ and time step $t$, we compute $|(S\hat{y}_b)_{i,t} - \hat{y}_{i,t}| / |\hat{y}_{i,t}|$, where $\hat{y}_b$ are the bottom-level mean forecasts and $S$ is the summing matrix. We report the mean and maximum over all aggregate series and time steps. This quantifies how much independently generated forecasts violate the hierarchical aggregation constraints before reconciliation.

To produce coherent probabilistic forecasts, we apply bottom-up reconciliation. We draw $K{=}100$ samples from the predicted quantile forecasts of the bottom-level series via inverse CDF interpolation: for each sample, a uniform random variable $u \sim \text{Uniform}(0,1)$ is drawn and mapped through the predicted quantile function to obtain a forecast value. These samples are then aggregated according to the summing matrix $S$ to produce coherent samples at all levels. Quantiles of the reconciled forecasts are computed from the resulting sample distribution.

\subsection{User guide}

\paragraph{When to use \ourmodel}
\ourmodel is designed for forecasting regularly-spaced, real-valued time series and can be used as a drop-in replacement for most forecasting methods --- it requires no task-specific training, producing forecasts directly from historical context.
It supports univariate forecasting, multivariate forecasting, and forecasting with time-varying covariates (both real-valued and categorical). The model is particularly suitable when little task-specific training data are available, when rapid deployment is required, or when related series and auxiliary covariates may provide signal beyond the target history alone. In our experiments, the same model generalizes across domains from retail to epidemiology without modification. Because \ourmodel operates in zero-shot mode, practitioners can quickly iterate on covariate selection and feature engineering without retraining. When task-specific data is available, fine-tuning can further improve performance. More broadly, \ourmodel can serve as a building block within larger forecasting systems that incorporate preprocessing, feature engineering, ensembling, or fine-tuning.

\paragraph{Data preparation}
\ourmodel supports several input formats.
For univariate forecasting, the simplest option is to provide an array of historical observations.
More generally, inputs can be organized as a long-format dataframe --- that is, a table where each row contains a series identifier, a timestamp, one or more target values, and optional covariates.
The user can provide an additional table containing the values of known future covariates over the forecast horizon.
The model assumes a fixed sampling frequency (e.g., daily or hourly), so irregularly spaced observations must be resampled before inference. Missing target or covariate values can be represented as \texttt{NaN}, and will be natively handled by the model.

\paragraph{Computational requirements}
Thanks to its small size (120M parameters), zero-shot inference with \ourmodel is feasible on standard CPU hardware such as a low-end laptop. On a mid-range GPU, the model processes 100 time series in under 1 second.
GPUs chiefly improve throughput for long contexts, large batches, or many jointly processed series. Fine-tuning remains possible on CPU, but a GPU offers significantly faster runtime.

\paragraph{Hyperparameters}
In zero-shot mode, the two hyperparameters are \texttt{cross\_learning} and \texttt{context\_length}. The \texttt{cross\_learning} mode is disabled by default to ensure independent predictions. When enabled, the model shares information across series in the same batch. This improves accuracy when the grouped series share relevant structure (e.g., related products, sensors on the same system), but may provide no benefit or even hurt performance when series are unrelated or exhibit very different dynamics. We recommend backtesting with and without \texttt{cross\_learning} before deployment. Increasing \texttt{context\_length} generally improves accuracy, provided the additional history remains informative, up to the model limit of 8,192 past time steps. Fine-tuning introduces additional hyperparameters, such as learning rate and batch size, that can be tuned for optimal performance.

\paragraph{Limitations}
Forecasting targets must be real-valued; categorical variables are supported only as covariates (via target~\citep{prokhorenkova2018catboost} or ordinal encoding), not as targets. Static covariates are currently not supported. The model supports up to 8,192 time steps of history; performance deteriorates with fewer than approximately 10 observations. For settings that require longer histories (e.g., hourly data with annual seasonality), fine-tuning with longer context or aggregating to a coarser frequency is recommended. For coarse-frequency data (monthly and below) with short histories, we recommend comparing against statistical baselines as they may be competitive.
Although \ourmodel was pretrained with a maximum group size of 32, we have observed cross-learning benefits extend to roughly 256 series in practice; beyond that, we recommend processing subgroups independently or fine-tuning on the full dataset.
The model most effectively captures lead-lag relationships within approximately 30 time steps; for longer-lag relationships, consider fine-tuning or downsampling to a coarser resolution.
\ourmodel outputs marginal predictive quantiles (0.01 to 0.99) for each future time step and dimension. It does not provide a joint predictive distribution and cannot natively generate sample trajectories with coherent dependence across dimensions and time.
Repeated forecasts at successive origins each require a full forward pass; the model cannot incrementally update predictions as new data arrives. For the vast majority of use cases --- including batch forecasting and offline backtesting --- inference is fast (under 1 second per 100 series on a mid-range GPU), but in real-time streaming settings with strict latency requirements, recursive methods may be preferable.

\clearpage

\section{Supplementary tables and figures}\label{app:supp}

\renewcommand{\thetable}{S\arabic{table}}
\setcounter{table}{0}
\renewcommand{\thefigure}{S\arabic{figure}}
\setcounter{figure}{0}

\begin{table}[ht]
    \centering
    \resizebox{\textwidth}{!}{%
    \begin{tabular}{llrll}
    \toprule
    \textbf{Dataset Name} & \textbf{Frequencies} & \textbf{\# Time Series} & \textbf{Domain} & \textbf{Source} \\
    \midrule
    Electricity & 15min, 1H, 1W, 1D & 370 & Energy & \citet{godahewa2021monash} \\
    KDD Cup (2018) & 1H, 1D & 270 & Nature & \citet{godahewa2021monash} \\
    M4 (Daily) & 1D & 4227 & Various & \citet{makridakis2020m4} \\
    M4 (Hourly) & 1H & 414 & Various & \citet{makridakis2020m4} \\
    M4 (Monthly) & 1M & 48000 & Various & \citet{makridakis2020m4} \\
    M4 (Weekly) & 1W & 359 & Various & \citet{makridakis2020m4} \\
    Mexico City Bikes & 1H, 1D, 1W & 494 & Transport & \citet{ansari2024chronos} \\
    Pedestrian Counts & 1H, 1D, 1W & 66 & Transport & \citet{godahewa2021monash} \\
    Solar & 5min, 10min, 1H & 5166 & Energy & \citet{ansari2024chronos} \\
    Taxi & 30min, 1H & 2428 & Transport & \citet{salinas2019high} \\
    Uber TLC & 1H, 1D & 262 & Transport & \citet{fivethirtyeight_uber_tlc_foil_response} \\
    USHCN & 1D, 1W & 225280 & Nature & \citet{ansari2024chronos} \\
    Weatherbench & 1H, 1D, 1W & 225280 & Nature & \citet{rasp2020weatherbench} \\
    Wiki & 1H, 1D, 1W & 100000 & Web & \citet{ansari2024chronos} \\
    Wind Farms & 1H, 1D & 337 & Energy & \citet{godahewa2021monash} \\
    Temperature-Rain & 1D & 32072 & Nature & \citet{godahewa2021monash} \\
    London Smart Meters & 30min, 1D & 5560 & Energy & \citet{godahewa2021monash} \\
    Alibaba Cluster Trace (2018) & 5min, 1H & 100000 & Cloud Ops & \citet{woo2023pushing} \\
    Azure VM Traces (2017) & 5min, 1H & 100000 & Cloud Ops & \citet{woo2023pushing} \\
    Borg Cluster Data (2011) & 5min, 1H & 100000 & Cloud Ops & \citet{woo2023pushing} \\
    LargeST (2017) & 1H, 1D & 8196 & Transport & \citet{liu2023largest} \\
    Q-Traffic & 15min, 1H & 45148 & Transport & \citet{jiang2023libcity} \\
    Buildings 900K & 1H, 1D & 100000 & Energy & \citet{emami2023buildingsbench} \\
    \bottomrule
    \end{tabular}%
    }
    \caption{Real univariate datasets used for pretraining \ourmodel.}
    \label{tab:real-uni-datasets}
\end{table}

\begin{table}[htb]
\centering
\resizebox{\textwidth}{!}{%

\begin{tabular}{p{7.5cm}cc}
\toprule
\textbf{Task Type} & \textbf{Group IDs} $\vg$ & \textbf{Future Inputs} $\mW$ \\
\midrule
Univariate Forecasting \par (\emph{3 independent series}) & 
$\vg = (1, 2, 3)$ & 
$\mW = \begin{bmatrix}
\ast & \dots & \ast \\
\ast & \dots & \ast \\
\ast & \dots & \ast
\end{bmatrix} \in \mathbb{R}^{3 \times H}$ \\
\cmidrule{1-3}
Multivariate Forecasting \par (\emph{3 targets}) & 
$\vg = (1, 1, 1)$ & 
$\mW = \begin{bmatrix}
\ast & \dots & \ast \\
\ast & \dots & \ast \\
\ast & \dots & \ast
\end{bmatrix} \in \mathbb{R}^{3 \times H}$ \\
\cmidrule{1-3}
Forecasting with Covariates \par (\emph{1 target, 1 past-only covariate, 2 known covariates}) & 
$\vg = (1, 1, 1, 1)$ & 
$
\mW = \begin{bmatrix}
\ast & \dots & \ast \\
\ast & \dots & \ast \\
x_{T+1,3} & \dots & x_{T+H,3} \\
x_{T+1,4} & \dots & x_{T+H,4}
\end{bmatrix} \in \mathbb{R}^{4 \times H}
$ \\
\bottomrule
\end{tabular}%

}
    \caption{Diverse forecasting tasks can be solved by specifying group IDs and future inputs appropriately. Here, $\vg$ and $\mW$ denote the group IDs and future values provided to the model. Future inputs for targets and past-only covariates are masked as missing values, denoted as $\ast$. The examples use fixed numbers of variates for clarity, but \ourmodel can handle arbitrary dimensions.}
    \label{tab:forecasting-tasks}
\end{table}

\begin{table}[h]
\centering
\resizebox{\textwidth}{!}{%

\begin{tabular}{lccc}
\toprule
& \textbf{Chronos-2} & \textbf{Chronos-Bolt} & \textbf{Chronos} \\
\midrule
Input Type & Patches & Patches & Individual Observations \\
Output Type & Multi-step Quantile Forecast & Multi-step Quantile Forecast & Autoregressive Sampling \\
Loss Function & Quantile Loss & Quantile Loss & Cross Entropy Loss \\
\midrule
Univariate Forecasting & \checkmark & \checkmark & \checkmark \\
Cross Learning & \checkmark & -- & -- \\
Multivariate Forecasting & \checkmark & -- & -- \\
Past-only Covariates & \checkmark & -- & -- \\
Known-future Covariates & \checkmark & -- & -- \\
\midrule
Max.\ Context Length & 8192 & 2048 & 512 \\
Max.\ Prediction Length & 1024+ & 64+ & 64+ \\
\bottomrule
\end{tabular}%

}
\caption{\textbf{Comparison of \ourmodel{} with previous versions of the Chronos family.}}
\label{tab:chronos-comparison}
\end{table}

\begin{table}[h]
    \centering
    \resizebox{\textwidth}{!}{
    \begin{tabular}{lrlrlrr}
\toprule
\textbf{Model} & \textbf{Skill Score (\%)} & \textbf{Skill Score 95\% CI} & \textbf{Avg. Win Rate (\%)} & \textbf{Win Rate 95\% CI} & \textbf{Median runtime (s)} & \textbf{\#Failures} \\
\midrule
\rowcolor{AccentColorLight} Chronos-2 & \bfseries 35.5 & (31.6, 39.5) & \bfseries 88.8 & (86.2, 91.2) & 0.8 & 0 \\
TiRex & 30.0 & (26.4, 33.8) & 79.8 & (76.8, 82.8) & 0.2 & 0 \\
TimesFM-2.5 & 30.3 & (26.7, 34.2) & 79.2 & (75.5, 82.8) & 1.9 & 0 \\
Toto-1.0 & 28.2 & (23.9, 32.7) & 69.8 & (64.9, 74.3) & 22.1 & 0 \\
Moirai-2.0 & 27.3 & (23.3, 31.3) & 65.5 & (61.1, 69.2) & 0.3 & 0 \\
Chronos-Bolt & 26.5 & (22.5, 30.5) & 65.1 & (61.2, 68.7) & 0.2 & 0 \\
TabPFN-TS & 27.6 & (23.0, 32.0) & 61.7 & (56.5, 66.7) & 88.9 & 2 \\
CatBoost & 23.7 & (18.9, 28.2) & 54.9 & (49.3, 60.2) & 29.3 & 0 \\
LightGBM & 21.7 & (16.2, 26.4) & 51.7 & (46.3, 57.1) & 3.0 & 0 \\
Moirai-1.1 & 20.1 & (15.3, 24.3) & 50.4 & (47.0, 53.8) & 6.9 & 0 \\
TFT & 20.5 & (15.7, 25.5) & 48.9 & (43.4, 54.2) & 827.2 & 0 \\
Stat-Ens & 15.7 & (11.4, 19.8) & 47.2 & (41.9, 53.1) & 148.6 & 11 \\
PatchTST & 18.4 & (13.4, 23.6) & 43.3 & (38.1, 48.4) & 662.1 & 0 \\
DeepAR & 17.5 & (13.6, 21.6) & 41.1 & (36.6, 45.6) & 1113.5 & 3 \\
Granite-TTM-R2 & 8.3 & (1.7, 13.8) & 38.0 & (33.2, 42.9) & 0.1 & 0 \\
AutoARIMA & 11.2 & (6.9, 15.4) & 37.5 & (33.1, 42.2) & 19.5 & 10 \\
AutoTheta & 11.0 & (7.1, 14.5) & 35.0 & (30.8, 39.2) & 3.2 & 0 \\
AutoETS & 2.3 & (-6.8, 9.4) & 34.5 & (28.5, 40.5) & 3.5 & 3 \\
TimePFN & -4.0 & (-11.0, 2.3) & 20.6 & (17.1, 24.2) & 0.1 & 0 \\
SeasonalNaive & 0.0 & (0.0, 0.0) & 20.0 & (16.2, 24.0) & 0.5 & 0 \\
Naive & -16.7 & (-29.2, -7.5) & 17.0 & (13.4, 21.0) & 0.5 & 0 \\
\bottomrule
\end{tabular}

    }
    \caption{\textbf{Point forecasting results on \fevbench.} The average win rate and skill score are computed with respect to the mean absolute scaled error (MASE) metric. Higher values are better for both. Scores for models with failures on specific tasks were imputed with the baseline (Seasonal Naive) score, as described in \citet{shchur2025fev}.}
    \vspace{-1em}
    \label{tab:fev-results-mase}
\end{table}

\begin{table}[h]
    \centering
    \resizebox{\textwidth}{!}{
    \begin{tabular}{lrlrlrr}
\toprule
\textbf{Model} & \textbf{Skill Score (\%)} & \textbf{Skill Score 95\% CI} & \textbf{Avg. Win Rate (\%)} & \textbf{Win Rate 95\% CI} & \textbf{Median runtime (s)} & \textbf{\#Failures} \\
\midrule
\rowcolor{AccentColorLight} Chronos-2 & \bfseries 47.3 & (42.9, 52.2) & \bfseries 92.4 & (90.5, 94.1) & 0.8 & 0 \\
TiRex & 42.6 & (38.1, 47.5) & 85.2 & (82.6, 87.6) & 0.2 & 0 \\
TimesFM-2.5 & 42.3 & (37.9, 47.4) & 81.5 & (78.2, 84.6) & 1.9 & 0 \\
Toto-1.0 & 40.7 & (35.8, 46.3) & 75.1 & (71.1, 78.6) & 22.1 & 0 \\
Moirai-2.0 & 39.3 & (34.4, 44.5) & 70.2 & (66.6, 73.1) & 0.3 & 0 \\
Chronos-Bolt & 38.9 & (34.2, 44.1) & 69.8 & (66.9, 72.5) & 0.2 & 0 \\
TabPFN-TS & 39.6 & (34.6, 44.7) & 67.5 & (63.0, 72.0) & 88.9 & 2 \\
Moirai-1.1 & 34.2 & (28.9, 39.5) & 60.6 & (57.5, 63.7) & 6.9 & 0 \\
TFT & 32.2 & (26.1, 38.7) & 54.8 & (50.0, 59.6) & 827.2 & 0 \\
PatchTST & 30.1 & (23.7, 36.9) & 48.5 & (44.1, 53.2) & 662.1 & 0 \\
DeepAR & 29.1 & (23.2, 35.7) & 47.6 & (43.4, 52.0) & 1113.5 & 3 \\
Stat-Ens & 20.2 & (14.3, 24.8) & 45.5 & (40.7, 50.6) & 148.6 & 11 \\
AutoARIMA & 20.6 & (14.8, 25.9) & 42.8 & (38.8, 46.9) & 19.5 & 10 \\
CatBoost & 23.0 & (16.6, 29.7) & 37.3 & (33.0, 41.4) & 29.3 & 0 \\
LightGBM & 21.0 & (14.2, 27.9) & 34.3 & (30.2, 38.3) & 3.0 & 0 \\
AutoETS & -26.8 & (-64.8, -0.6) & 34.1 & (28.3, 40.0) & 3.5 & 3 \\
AutoTheta & 5.5 & (-3.2, 12.3) & 29.5 & (25.1, 33.6) & 3.2 & 0 \\
Granite-TTM-R2 & 7.4 & (-1.0, 15.6) & 26.0 & (22.9, 29.3) & 0.1 & 0 \\
SeasonalNaive & 0.0 & (0.0, 0.0) & 20.5 & (17.8, 23.4) & 0.5 & 0 \\
TimePFN & -4.9 & (-13.6, 3.7) & 15.4 & (13.0, 18.0) & 0.1 & 0 \\
Naive & -45.4 & (-62.9, -30.3) & 11.2 & (8.4, 14.1) & 0.5 & 0 \\
\bottomrule
\end{tabular}

    }
    \caption{\textbf{Probabilistic forecasting results on \fevbench.} The average win rate and skill score are computed with respect to the scaled quantile loss (SQL) metric. Higher values are better for both. Scores for models with failures on specific tasks were imputed with the baseline (Seasonal Naive) score, as described in \citet{shchur2025fev}.}
    \vspace{-1em}
    \label{tab:fev-results-sql}
\end{table}

\begin{table}[h]
    \centering
    \subfloat[Point forecasting (MASE)]{
    \resizebox{0.46\textwidth}{!}{
    \begin{tabular}{lll}
\toprule
\textbf{Model} & \textbf{p-value} & \textbf{Sig.} \\
\midrule
TimesFM-2.5 & $1.68 \times 10^{-7}$ & *** \\
TiRex & $1.31 \times 10^{-7}$ & *** \\
Toto-1.0 & $1.60 \times 10^{-9}$ & *** \\
TabPFN-TS & $1.05 \times 10^{-10}$ & *** \\
CatBoost & $4.87 \times 10^{-12}$ & *** \\
LightGBM & $1.63 \times 10^{-12}$ & *** \\
Moirai-2.0 & $1.36 \times 10^{-13}$ & *** \\
Stat-Ens & $4.44 \times 10^{-14}$ & *** \\
TFT & $9.63 \times 10^{-15}$ & *** \\
Chronos-Bolt & $9.63 \times 10^{-15}$ & *** \\
Moirai-1.1 & $7.18 \times 10^{-15}$ & *** \\
AutoETS & $5.29 \times 10^{-15}$ & *** \\
DeepAR & $2.07 \times 10^{-15}$ & *** \\
PatchTST & $1.58 \times 10^{-15}$ & *** \\
AutoARIMA & $1.35 \times 10^{-15}$ & *** \\
Granite-TTM-R2 & $7.20 \times 10^{-16}$ & *** \\
AutoTheta & $4.94 \times 10^{-16}$ & *** \\
TimePFN & $1.43 \times 10^{-16}$ & *** \\
Naive & $5.48 \times 10^{-17}$ & *** \\
SeasonalNaive & $4.96 \times 10^{-17}$ & *** \\
\bottomrule
\end{tabular}

    }
    }
    \quad
    \subfloat[Probabilistic forecasting (SQL)]{
    \resizebox{0.46\textwidth}{!}{
    \begin{tabular}{lll}
\toprule
\textbf{Model} & \textbf{p-value} & \textbf{Sig.} \\
\midrule
TiRex & $1.65 \times 10^{-7}$ & *** \\
TimesFM-2.5 & $1.24 \times 10^{-8}$ & *** \\
Toto-1.0 & $1.27 \times 10^{-10}$ & *** \\
TabPFN-TS & $3.27 \times 10^{-12}$ & *** \\
Moirai-2.0 & $1.30 \times 10^{-14}$ & *** \\
Moirai-1.1 & $3.13 \times 10^{-15}$ & *** \\
Chronos-Bolt & $2.84 \times 10^{-15}$ & *** \\
TFT & $1.17 \times 10^{-15}$ & *** \\
Stat-Ens & $6.25 \times 10^{-16}$ & *** \\
AutoARIMA & $2.74 \times 10^{-16}$ & *** \\
PatchTST & $2.68 \times 10^{-16}$ & *** \\
DeepAR & $2.61 \times 10^{-16}$ & *** \\
AutoETS & $2.10 \times 10^{-16}$ & *** \\
AutoTheta & $6.14 \times 10^{-17}$ & *** \\
CatBoost & $5.34 \times 10^{-17}$ & *** \\
LightGBM & $5.21 \times 10^{-17}$ & *** \\
Granite-TTM-R2 & $5.21 \times 10^{-17}$ & *** \\
SeasonalNaive & $3.90 \times 10^{-17}$ & *** \\
TimePFN & $3.90 \times 10^{-17}$ & *** \\
Naive & $3.90 \times 10^{-17}$ & *** \\
\bottomrule
\end{tabular}

    }
    }
    
    \caption{Holm-Šidák (HS) adjusted $p$ values of the one-sided Wilcoxon signed rank test comparing \ourmodel against all baselines on \fevbench. ***$p < 0.001$, **$p < 0.01$, *$p < 0.05$.}
    \vspace{-1em}
    \label{tab:fev-bench-pval}
\end{table}

\begin{table}[h]
    \centering
    \resizebox{\textwidth}{!}{
    \begin{tabular}{lrlrl}
\toprule
\textbf{Model} & \textbf{Skill Score (\%)} & \textbf{Skill Score 95\% CI} & \textbf{Avg. Win Rate (\%)} & \textbf{Win Rate 95\% CI} \\
\midrule
\rowcolor{AccentColorLight} Chronos-2 & \bfseries 26.5 & (19.9, 33.7) & \bfseries 79.5 & (70.6, 87.7) \\
TimesFM-2.5 & 23.3 & (17.4, 29.8) & 72.1 & (64.7, 79.5) \\
TiRex & 22.2 & (15.7, 28.9) & 67.7 & (60.5, 73.3) \\
Stat-Ens & 19.6 & (12.4, 27.4) & 64.9 & (52.8, 76.8) \\
Toto-1.0 & 22.3 & (16.2, 28.5) & 60.7 & (51.9, 70.6) \\
DeepAR & 18.4 & (12.0, 24.9) & 57.5 & (46.4, 67.7) \\
Moirai-2.0 & 19.8 & (13.2, 26.9) & 53.6 & (43.9, 64.7) \\
TFT & 19.2 & (12.2, 26.1) & 53.5 & (41.7, 64.7) \\
Chronos-Bolt & 20.4 & (14.7, 26.7) & 52.3 & (45.4, 58.8) \\
PatchTST & 17.7 & (10.9, 24.5) & 51.7 & (41.0, 62.0) \\
Moirai-1.1 & 17.8 & (11.0, 25.1) & 48.6 & (41.5, 56.1) \\
AutoARIMA & 16.0 & (8.5, 24.2) & 48.4 & (36.5, 59.0) \\
TabPFN-TS & 10.5 & (0.7, 19.2) & 45.6 & (35.9, 54.7) \\
Seasonal Naive & 0.0 & (0.0, 0.0) & 19.1 & (13.6, 25.3) \\
Granite-TTM-R2 & -7.0 & (-20.4, 5.3) & 16.8 & (10.6, 24.9) \\
TimePFN & -25.0 & (-45.7, -7.6) & 7.9 & (4.4, 11.6) \\
\bottomrule
\end{tabular}

    }
    \caption{\textbf{Point forecasting results on \chronosbenchii.} The average win rate and skill score are computed with respect to the mean absolute scaled error (MASE) metric. Higher values are better for both. Scores for models with failures on specific tasks were imputed with the baseline (Seasonal Naive) score, as described in \citet{ansari2024chronos}.}
    \vspace{-1em}
    \label{tab:zs-mase}
\end{table}

\begin{table}[h]
    \centering
    \resizebox{\textwidth}{!}{
    \begin{tabular}{lrlrl}
\toprule
\textbf{Model} & \textbf{Skill Score (\%)} & \textbf{Skill Score 95\% CI} & \textbf{Avg. Win Rate (\%)} & \textbf{Win Rate 95\% CI} \\
\midrule
\rowcolor{AccentColorLight} Chronos-2 & \bfseries 46.6 & (36.3, 56.4) & \bfseries 79.5 & (70.4, 87.7) \\
TiRex & 41.7 & (31.0, 52.1) & 73.8 & (65.9, 80.2) \\
TimesFM-2.5 & 42.4 & (31.7, 52.8) & 72.3 & (65.7, 78.8) \\
Toto-1.0 & 41.9 & (30.6, 52.4) & 64.4 & (55.6, 73.3) \\
Moirai-2.0 & 40.9 & (28.9, 51.7) & 61.0 & (50.4, 71.6) \\
TFT & 36.2 & (24.9, 46.9) & 57.9 & (46.8, 69.3) \\
Chronos-Bolt & 39.3 & (29.2, 49.3) & 56.5 & (50.4, 62.5) \\
DeepAR & 29.9 & (9.9, 44.5) & 56.3 & (46.0, 66.7) \\
PatchTST & 32.5 & (15.8, 45.0) & 54.0 & (43.2, 64.8) \\
Stat-Ens & 33.0 & (22.0, 44.0) & 51.9 & (42.2, 61.0) \\
TabPFN-TS & 32.6 & (21.6, 44.2) & 49.3 & (40.7, 58.0) \\
Moirai-1.1 & 37.7 & (25.7, 49.1) & 48.4 & (41.0, 56.3) \\
AutoARIMA & 30.4 & (18.2, 42.6) & 40.0 & (31.8, 50.1) \\
Seasonal Naive & 0.0 & (0.0, 0.0) & 17.7 & (12.0, 23.5) \\
Granite-TTM-R2 & -6.2 & (-28.9, 15.4) & 9.4 & (6.7, 12.3) \\
TimePFN & -19.0 & (-49.4, 8.2) & 7.7 & (3.0, 15.6) \\
\bottomrule
\end{tabular}

    }
    \caption{\textbf{Probabilistic forecasting results on \chronosbenchii.} The average win rate and skill score are computed with respect to the weighted quantile loss (WQL) metric. Higher values are better for both. Scores for models with failures on specific tasks were imputed with the baseline (Seasonal Naive) score, as described in \citet{ansari2024chronos}.}
    \vspace{-1em}
    \label{tab:zs-wql}
\end{table}

\begin{table}[h]
    \centering
    \subfloat[Point forecasting (MASE)]{
    \resizebox{0.46\textwidth}{!}{
    \begin{tabular}{lll}
\toprule
\textbf{Model} & \textbf{p-value} & \textbf{Sig.} \\
\midrule
DeepAR & 0.045 & * \\
Stat-Ens & 0.045 & * \\
TFT & 0.045 & * \\
TimesFM-2.5 & 0.045 & * \\
TiRex & 0.035 & * \\
Moirai-2.0 & 0.031 & * \\
Toto-1.0 & 0.028 & * \\
PatchTST & 0.026 & * \\
Chronos-Bolt & 0.0027 & ** \\
AutoARIMA & 0.0021 & ** \\
Moirai-1.1 & $5.28 \times 10^{-4}$ & *** \\
TabPFN-TS & $5.02 \times 10^{-4}$ & *** \\
Seasonal Naive & $2.42 \times 10^{-6}$ & *** \\
Granite-TTM-R2 & $5.22 \times 10^{-7}$ & *** \\
TimePFN & $1.12 \times 10^{-7}$ & *** \\
\bottomrule
\end{tabular}

    }
    }
    \quad
    \subfloat[Probabilistic forecasting (WQL)]{
    \resizebox{0.46\textwidth}{!}{
    \begin{tabular}{lll}
\toprule
\textbf{Model} & \textbf{p-value} & \textbf{Sig.} \\
\midrule
DeepAR & 0.046 & * \\
PatchTST & 0.046 & * \\
TFT & 0.046 & * \\
TimesFM-2.5 & 0.046 & * \\
Toto-1.0 & 0.046 & * \\
Moirai-2.0 & 0.033 & * \\
TiRex & 0.028 & * \\
Stat-Ens & 0.0050 & ** \\
Chronos-Bolt & 0.0019 & ** \\
TabPFN-TS & 0.0017 & ** \\
Moirai-1.1 & $6.89 \times 10^{-4}$ & *** \\
AutoARIMA & $4.37 \times 10^{-4}$ & *** \\
Seasonal Naive & $4.84 \times 10^{-7}$ & *** \\
Granite-TTM-R2 & $1.12 \times 10^{-7}$ & *** \\
TimePFN & $1.12 \times 10^{-7}$ & *** \\
\bottomrule
\end{tabular}

    }
    }
    
    \caption{Holm-Šidák (HS) adjusted $p$ values of the one-sided Wilcoxon signed rank test comparing \ourmodel against all baselines on \chronosbenchii. ***$p < 0.001$, **$p < 0.01$, *$p < 0.05$.}
    \vspace{-1em}
    \label{tab:zs-pval}
\end{table}

\begin{table}[h]
    \centering
    \resizebox{\textwidth}{!}{
    \begin{tabular}{lrlrl}
\toprule
\textbf{Model} & \textbf{Skill Score (\%)} & \textbf{Skill Score 95\% CI} & \textbf{Avg. Win Rate (\%)} & \textbf{Win Rate 95\% CI} \\
\midrule
\rowcolor{AccentColorLight} Chronos-2 & \bfseries 30.2 & (26.5, 34.5) & \bfseries 87.8 & (84.4, 90.7) \\
TimesFM-2.5 & 29.5 & (25.5, 34.2) & 84.4 & (81.5, 87.3) \\
TiRex & 27.6 & (23.8, 31.8) & 79.0 & (75.1, 82.4) \\
Moirai-2.0 & 27.2 & (23.3, 31.5) & 74.5 & (71.0, 77.8) \\
Toto-1.0 & 25.2 & (21.7, 29.1) & 72.9 & (69.7, 76.1) \\
Chronos-Bolt & 19.2 & (11.2, 25.9) & 68.6 & (64.0, 73.1) \\
TabPFN-TS & 16.6 & (10.0, 22.1) & 59.5 & (55.9, 63.1) \\
Moirai-1.1 & 9.9 & (0.1, 17.4) & 53.0 & (47.9, 57.5) \\
PatchTST & 15.1 & (9.4, 20.9) & 49.9 & (45.9, 54.4) \\
TFT & 8.5 & (-0.4, 16.1) & 41.7 & (37.1, 46.5) \\
Stat-Ens & -52.6 & (-111.6, -12.2) & 36.2 & (31.6, 40.9) \\
Granite-TTM-R2 & -2.0 & (-10.5, 5.4) & 34.9 & (30.6, 39.6) \\
AutoARIMA & -7.4 & (-42.3, 4.8) & 33.9 & (29.4, 38.6) \\
Seasonal Naive & 0.0 & (0.0, 0.0) & 28.8 & (25.1, 33.1) \\
AutoETS & -21.2 & (-34.8, -10.3) & 26.2 & (21.3, 31.2) \\
AutoTheta & -9.0 & (-17.4, -1.0) & 25.0 & (20.4, 29.7) \\
DeepAR & -34.3 & (-58.1, -16.8) & 24.0 & (19.5, 28.8) \\
TimePFN & -18.4 & (-29.2, -8.5) & 19.7 & (16.6, 23.1) \\
\bottomrule
\end{tabular}

    }
    \caption{\textbf{Point forecasting results on \gifteval.} The average win rate and skill score are computed with respect to the mean absolute scaled error (MASE) metric. Higher values are better for both. Scores for models with failures on specific tasks were imputed with the baseline (Seasonal Naive) score, as described in \citet{aksu2024gifteval}.}
    \vspace{-1em}
    \label{tab:gift-eval-mase}
\end{table}

\begin{table}[h]
    \centering
    \resizebox{\textwidth}{!}{
    \begin{tabular}{lrlrl}
\toprule
\textbf{Model} & \textbf{Skill Score (\%)} & \textbf{Skill Score 95\% CI} & \textbf{Avg. Win Rate (\%)} & \textbf{Win Rate 95\% CI} \\
\midrule
\rowcolor{AccentColorLight} Chronos-2 & \bfseries 51.4 & (47.0, 55.3) & \bfseries 85.9 & (82.7, 88.9) \\
TimesFM-2.5 & 51.0 & (46.4, 55.3) & 83.4 & (79.7, 86.4) \\
TiRex & 50.2 & (45.7, 54.4) & 81.6 & (78.1, 84.7) \\
Toto-1.0 & 48.6 & (43.9, 52.9) & 76.3 & (73.1, 79.4) \\
Moirai-2.0 & 48.4 & (43.7, 52.8) & 72.8 & (69.3, 76.0) \\
TabPFN-TS & 43.1 & (37.9, 48.0) & 65.2 & (61.6, 68.6) \\
Chronos-Bolt & 42.6 & (36.5, 48.1) & 64.2 & (59.8, 68.8) \\
Moirai-1.1 & 39.0 & (32.6, 44.9) & 59.4 & (54.7, 63.4) \\
PatchTST & 41.3 & (36.3, 46.2) & 57.1 & (53.8, 60.5) \\
TFT & 39.5 & (34.0, 45.1) & 54.4 & (50.5, 59.0) \\
DeepAR & 14.7 & (1.9, 24.9) & 34.6 & (29.7, 39.2) \\
Stat-Ens & 10.3 & (1.9, 18.7) & 32.2 & (28.2, 36.1) \\
AutoARIMA & 8.8 & (-1.1, 15.8) & 28.5 & (24.9, 32.2) \\
Granite-TTM-R2 & 12.7 & (3.2, 20.8) & 25.6 & (22.6, 28.7) \\
Seasonal Naive & 0.0 & (0.0, 0.0) & 22.2 & (19.4, 25.7) \\
AutoTheta & -24.4 & (-41.8, -7.5) & 19.9 & (15.6, 24.7) \\
AutoETS & -648.9 & (-3747.6, -94.4) & 18.7 & (14.7, 23.1) \\
TimePFN & 0.3 & (-10.3, 9.6) & 18.0 & (15.6, 20.5) \\
\bottomrule
\end{tabular}

    }
    \caption{\textbf{Probabilistic forecasting results on \gifteval.} The average win rate and skill score are computed with respect to the weighted quantile loss (WQL) metric. Higher values are better for both. Scores for models with failures on specific tasks were imputed with the baseline (Seasonal Naive) score, as described in \citet{aksu2024gifteval}.}
    \vspace{-1em}
    \label{tab:gift-eval-wql}
\end{table}

\begin{table}[h]
    \centering
    \subfloat[Point forecasting (MASE)]{
    \resizebox{0.46\textwidth}{!}{
    \begin{tabular}{lll}
\toprule
\textbf{Model} & \textbf{p-value} & \textbf{Sig.} \\
\midrule
TimesFM-2.5 & 0.097 &  \\
TiRex & $3.76 \times 10^{-5}$ & *** \\
Moirai-2.0 & $6.38 \times 10^{-7}$ & *** \\
Chronos-Bolt & $4.04 \times 10^{-7}$ & *** \\
Toto-1.0 & $1.09 \times 10^{-8}$ & *** \\
Moirai-1.1 & $1.02 \times 10^{-11}$ & *** \\
TabPFN-TS & $6.09 \times 10^{-13}$ & *** \\
TFT & $1.02 \times 10^{-14}$ & *** \\
AutoARIMA & $2.49 \times 10^{-15}$ & *** \\
PatchTST & $2.49 \times 10^{-15}$ & *** \\
Stat-Ens & $1.38 \times 10^{-15}$ & *** \\
AutoETS & $4.11 \times 10^{-16}$ & *** \\
Granite-TTM-R2 & $3.59 \times 10^{-16}$ & *** \\
Seasonal Naive & $2.16 \times 10^{-16}$ & *** \\
DeepAR & $1.98 \times 10^{-16}$ & *** \\
AutoTheta & $1.87 \times 10^{-16}$ & *** \\
TimePFN & $1.04 \times 10^{-16}$ & *** \\
\bottomrule
\end{tabular}

    }
    }
    \quad
    \subfloat[Probabilistic forecasting (WQL)]{
    \resizebox{0.46\textwidth}{!}{
    \begin{tabular}{lll}
\toprule
\textbf{Model} & \textbf{p-value} & \textbf{Sig.} \\
\midrule
TimesFM-2.5 & 0.22 &  \\
TiRex & 0.0090 & ** \\
Toto-1.0 & $3.80 \times 10^{-5}$ & *** \\
Moirai-2.0 & $3.50 \times 10^{-6}$ & *** \\
Chronos-Bolt & $1.43 \times 10^{-7}$ & *** \\
Moirai-1.1 & $8.80 \times 10^{-9}$ & *** \\
TabPFN-TS & $1.83 \times 10^{-9}$ & *** \\
TFT & $5.08 \times 10^{-12}$ & *** \\
PatchTST & $1.05 \times 10^{-12}$ & *** \\
DeepAR & $3.34 \times 10^{-15}$ & *** \\
AutoTheta & $1.70 \times 10^{-16}$ & *** \\
Stat-Ens & $1.60 \times 10^{-16}$ & *** \\
Granite-TTM-R2 & $1.60 \times 10^{-16}$ & *** \\
TimePFN & $1.60 \times 10^{-16}$ & *** \\
AutoARIMA & $1.60 \times 10^{-16}$ & *** \\
AutoETS & $1.55 \times 10^{-16}$ & *** \\
Seasonal Naive & $1.04 \times 10^{-16}$ & *** \\
\bottomrule
\end{tabular}

    }
    }
    
    \caption{Holm-Šidák (HS) adjusted $p$ values of the one-sided Wilcoxon signed rank test comparing \ourmodel against all baselines on \gifteval. ***$p < 0.001$, **$p < 0.01$, *$p < 0.05$.}
    \vspace{-1em}
    \label{tab:gift-pval}
\end{table}

\begin{figure}[htb]
    \centering
    \includegraphics[width=\textwidth]{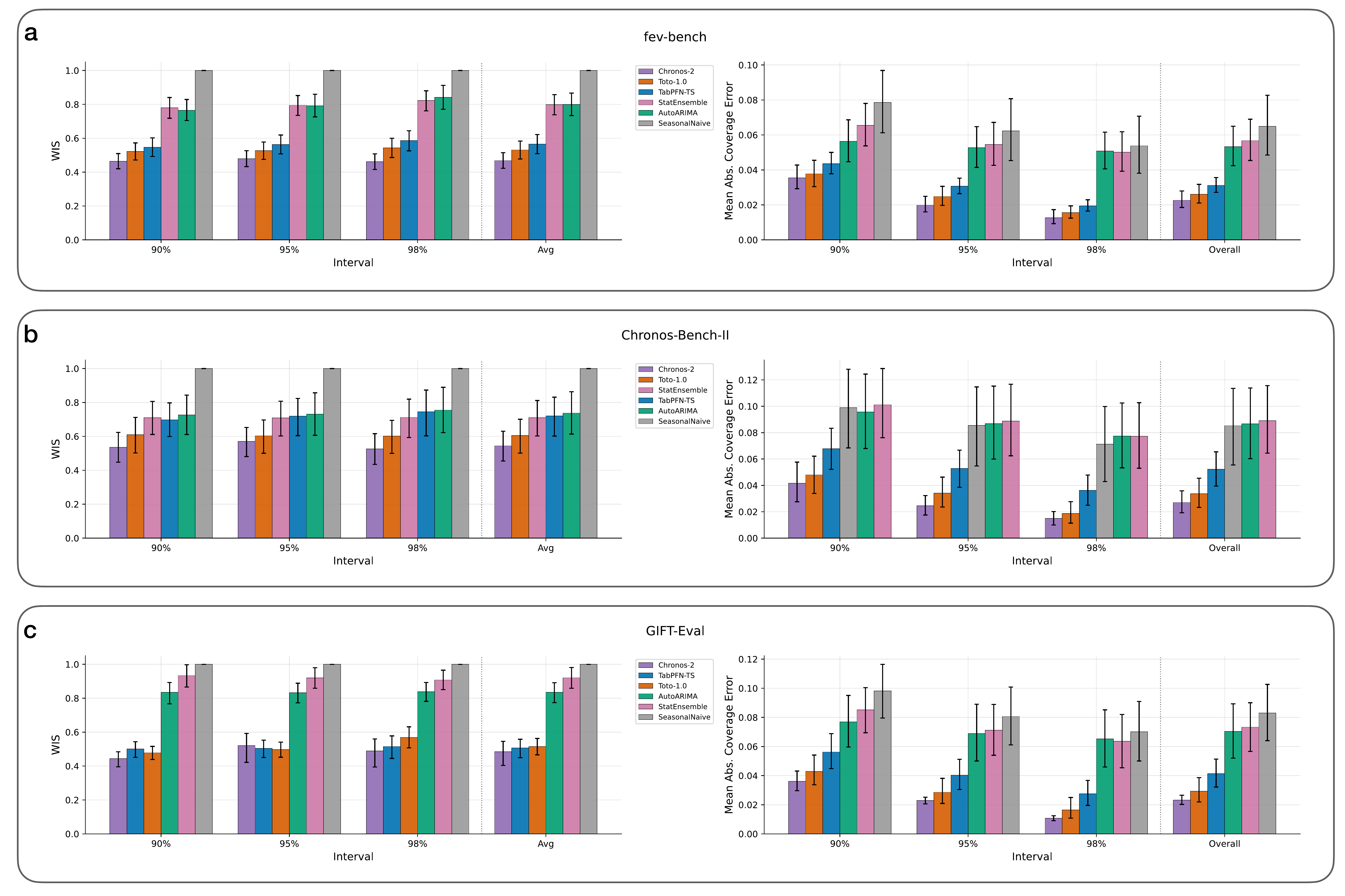}
    \caption{\textbf{Calibration of probabilistic forecasts across benchmarks} (\textit{best viewed in color on a screen}). Comparison of \ourmodel against baselines that support extreme quantiles on three central prediction intervals: 90\% $[0.05, 0.95]$, 95\% $[0.025, 0.975]$, and 98\% $[0.01, 0.99]$. Left panels show the geometric mean of the Weighted Interval Score (WIS), a strictly proper scoring rule that jointly assesses calibration and sharpness, relative to the Seasonal Naive baseline; lower values indicate better performance. Right panels show the Mean Absolute Coverage Error (MACE), which measures the discrepancy between nominal and empirical interval coverage; lower values indicate better calibration. Results are shown for (a) \fevbench, (b) \chronosbenchii, and (c) \gifteval. \ourmodel achieves the best calibration on 23 out of 24 metric-interval pairs across the three benchmarks.}
    \label{fig:calibration}
\end{figure}

\clearpage

\begin{table}
\centering
\resizebox{\textwidth}{!}{%

\begin{tabular}{lllrrrrrrrr}
\toprule
\textbf{Task} & \textbf{Domain} & \textbf{Freq.} & $H$ & $W$ & \textbf{Med. length} & \textbf{\# series} & \textbf{\# targets} & \textbf{\# past cov.} & \textbf{\# known cov.} & \textbf{\# static cov.} \\
\midrule
Australian Tourism & econ & Q & 8 & 2 & 36 & 89 & 1 & 0 & 0 & 0 \\
BizITObs - L2C & cloud & 5T & 288 & 20 & 31,968 & 1 & 7 & 0 & 0 & 0 \\
BizITObs - L2C & cloud & H & 24 & 20 & 2,664 & 1 & 7 & 0 & 0 & 0 \\
BOOMLET - 1062 & cloud & 5T & 288 & 20 & 16,384 & 1 & 21 & 0 & 0 & 0 \\
BOOMLET - 1209 & cloud & 5T & 288 & 20 & 16,384 & 1 & 53 & 0 & 0 & 0 \\
BOOMLET - 1225 & cloud & T & 60 & 20 & 16,384 & 1 & 49 & 0 & 0 & 0 \\
BOOMLET - 1230 & cloud & 5T & 288 & 20 & 16,384 & 1 & 23 & 0 & 0 & 0 \\
BOOMLET - 1282 & cloud & T & 60 & 20 & 16,384 & 1 & 35 & 0 & 0 & 0 \\
BOOMLET - 1487 & cloud & 5T & 288 & 20 & 16,384 & 1 & 54 & 0 & 0 & 0 \\
BOOMLET - 1631 & cloud & 30T & 96 & 20 & 10,463 & 1 & 40 & 0 & 0 & 0 \\
BOOMLET - 1676 & cloud & 30T & 96 & 20 & 10,463 & 1 & 100 & 0 & 0 & 0 \\
BOOMLET - 1855 & cloud & H & 24 & 20 & 5,231 & 1 & 52 & 0 & 0 & 0 \\
BOOMLET - 1975 & cloud & H & 24 & 20 & 5,231 & 1 & 75 & 0 & 0 & 0 \\
BOOMLET - 2187 & cloud & H & 24 & 20 & 5,231 & 1 & 100 & 0 & 0 & 0 \\
BOOMLET - 285 & cloud & T & 60 & 20 & 16,384 & 1 & 75 & 0 & 0 & 0 \\
BOOMLET - 619 & cloud & T & 60 & 20 & 16,384 & 1 & 52 & 0 & 0 & 0 \\
BOOMLET - 772 & cloud & T & 60 & 20 & 16,384 & 1 & 67 & 0 & 0 & 0 \\
BOOMLET - 963 & cloud & T & 60 & 20 & 16,384 & 1 & 28 & 0 & 0 & 0 \\
ECDC ILI & healthcare & W & 13 & 10 & 201 & 25 & 1 & 0 & 0 & 0 \\
ENTSO-e Load & energy & 15T & 96 & 20 & 175,292 & 6 & 1 & 0 & 3 & 0 \\
ENTSO-e Load & energy & 30T & 96 & 20 & 87,645 & 6 & 1 & 0 & 3 & 0 \\
ENTSO-e Load & energy & H & 168 & 20 & 43,822 & 6 & 1 & 0 & 3 & 0 \\
EPF-BE & energy & H & 24 & 20 & 52,416 & 1 & 1 & 0 & 2 & 0 \\
EPF-DE & energy & H & 24 & 20 & 52,416 & 1 & 1 & 0 & 2 & 0 \\
EPF-FR & energy & H & 24 & 20 & 52,416 & 1 & 1 & 0 & 2 & 0 \\
EPF-NP & energy & H & 24 & 20 & 52,416 & 1 & 1 & 0 & 2 & 0 \\
EPF-PJM & energy & H & 24 & 20 & 52,416 & 1 & 1 & 0 & 2 & 0 \\
ERCOT & energy & D & 28 & 20 & 6,452 & 8 & 1 & 0 & 0 & 0 \\
ERCOT & energy & H & 168 & 20 & 154,872 & 8 & 1 & 0 & 0 & 0 \\
ERCOT & energy & M & 12 & 15 & 211 & 8 & 1 & 0 & 0 & 0 \\
ERCOT & energy & W & 13 & 20 & 921 & 8 & 1 & 0 & 0 & 0 \\
ETT & energy & 15T & 96 & 20 & 69,680 & 2 & 7 & 0 & 0 & 0 \\
ETT & energy & D & 28 & 20 & 724 & 2 & 7 & 0 & 0 & 0 \\
ETT & energy & H & 168 & 20 & 17,420 & 2 & 7 & 0 & 0 & 0 \\
ETT & energy & W & 13 & 5 & 103 & 2 & 7 & 0 & 0 & 0 \\
Favorita Store Sales & retail & D & 28 & 10 & 1,688 & 1,579 & 1 & 1 & 2 & 6 \\
Favorita Store Sales & retail & M & 12 & 2 & 54 & 1,579 & 1 & 1 & 1 & 6 \\
Favorita Store Sales & retail & W & 13 & 10 & 240 & 1,579 & 1 & 1 & 1 & 6 \\
Favorita Transactions & retail & D & 28 & 10 & 1,688 & 51 & 1 & 1 & 1 & 5 \\
Favorita Transactions & retail & M & 12 & 2 & 54 & 51 & 1 & 1 & 0 & 5 \\
Favorita Transactions & retail & W & 13 & 10 & 240 & 51 & 1 & 1 & 0 & 5 \\
FRED-MD - CEE & econ & M & 12 & 20 & 798 & 1 & 3 & 4 & 0 & 0 \\
FRED-MD - Macro & econ & M & 12 & 20 & 798 & 1 & 51 & 0 & 0 & 0 \\
FRED-QD - CEE & econ & Q & 8 & 20 & 266 & 1 & 3 & 4 & 0 & 0 \\
FRED-QD - Macro & econ & Q & 8 & 20 & 266 & 1 & 51 & 0 & 0 & 0 \\
GFC12 & energy & H & 168 & 10 & 39,414 & 11 & 1 & 0 & 1 & 0 \\
GFC14 & energy & H & 168 & 20 & 17,520 & 1 & 1 & 0 & 1 & 0 \\
GFC17 & energy & H & 168 & 20 & 17,544 & 8 & 1 & 0 & 1 & 0 \\
GVAR & econ & Q & 8 & 10 & 178 & 33 & 6 & 3 & 0 & 0 \\
Hermes & retail & W & 52 & 1 & 261 & 10,000 & 1 & 0 & 1 & 2 \\
\bottomrule
\end{tabular}%

}
\caption{\textbf{Task statistics for \fevbench (Part 1/2).} $H$ denotes the forecast horizon, $W$ the number of rolling evaluation windows. Median length refers to the median number of time steps across series in a task.}
\label{tab:fev-task-stats}
\end{table}

\begin{table}
\centering
\resizebox{\textwidth}{!}{%

\begin{tabular}{lllrrrrrrrr}
\toprule
\textbf{Task} & \textbf{Domain} & \textbf{Freq.} & $H$ & $W$ & \textbf{Med. length} & \textbf{\# series} & \textbf{\# targets} & \textbf{\# past cov.} & \textbf{\# known cov.} & \textbf{\# static cov.} \\
\midrule
Hierarchical Sales & retail & D & 28 & 10 & 1,825 & 118 & 1 & 0 & 0 & 0 \\
Hierarchical Sales & retail & W & 13 & 10 & 260 & 118 & 1 & 0 & 0 & 0 \\
Hospital & healthcare & M & 12 & 4 & 84 & 767 & 1 & 0 & 0 & 0 \\
Hospital Admissions & healthcare & D & 28 & 20 & 1,731 & 8 & 1 & 0 & 0 & 0 \\
Hospital Admissions & healthcare & W & 13 & 16 & 246 & 8 & 1 & 0 & 0 & 0 \\
Jena Weather & nature & 10T & 144 & 20 & 52,704 & 1 & 21 & 0 & 0 & 0 \\
Jena Weather & nature & D & 28 & 11 & 366 & 1 & 21 & 0 & 0 & 0 \\
Jena Weather & nature & H & 24 & 20 & 8,784 & 1 & 21 & 0 & 0 & 0 \\
KDD Cup 2022 & energy & 10T & 288 & 10 & 35,279 & 134 & 1 & 9 & 0 & 0 \\
KDD Cup 2022 & energy & 30T & 96 & 10 & 11,758 & 134 & 1 & 9 & 0 & 0 \\
KDD Cup 2022 & energy & D & 14 & 10 & 243 & 134 & 1 & 9 & 0 & 0 \\
Loop Seattle & mobility & 5T & 288 & 10 & 105,120 & 323 & 1 & 0 & 0 & 0 \\
Loop Seattle & mobility & D & 28 & 10 & 365 & 323 & 1 & 0 & 0 & 0 \\
Loop Seattle & mobility & H & 168 & 10 & 8,760 & 323 & 1 & 0 & 0 & 0 \\
M-DENSE & mobility & D & 28 & 10 & 730 & 30 & 1 & 0 & 0 & 0 \\
M-DENSE & mobility & H & 168 & 10 & 17,520 & 30 & 1 & 0 & 0 & 0 \\
M5 & retail & D & 28 & 1 & 1,810 & 30,490 & 1 & 0 & 8 & 5 \\
M5 & retail & M & 12 & 1 & 58 & 30,490 & 1 & 0 & 8 & 5 \\
M5 & retail & W & 13 & 1 & 257 & 30,490 & 1 & 0 & 8 & 5 \\
Redset & cloud & 15T & 96 & 10 & 8,640 & 126 & 1 & 0 & 0 & 1 \\
Redset & cloud & 5T & 288 & 10 & 25,920 & 118 & 1 & 0 & 0 & 1 \\
Redset & cloud & H & 24 & 10 & 2,160 & 138 & 1 & 0 & 0 & 1 \\
Restaurant & retail & D & 28 & 8 & 296 & 817 & 1 & 0 & 0 & 4 \\
Rohlik Orders & retail & D & 61 & 5 & 1,197 & 7 & 1 & 9 & 4 & 0 \\
Rohlik Orders & retail & W & 8 & 5 & 170 & 7 & 1 & 9 & 4 & 0 \\
Rohlik Sales & retail & D & 14 & 1 & 1,046 & 5,390 & 1 & 1 & 13 & 7 \\
Rohlik Sales & retail & W & 8 & 1 & 150 & 5,243 & 1 & 1 & 13 & 7 \\
Rossmann & retail & D & 48 & 10 & 942 & 1,115 & 1 & 1 & 5 & 10 \\
Rossmann & retail & W & 13 & 8 & 133 & 1,115 & 1 & 1 & 4 & 10 \\
Solar & energy & D & 28 & 10 & 365 & 137 & 1 & 0 & 0 & 0 \\
Solar & energy & W & 13 & 1 & 52 & 137 & 1 & 0 & 0 & 0 \\
Solar with Weather & energy & 15T & 96 & 20 & 198,600 & 1 & 1 & 2 & 7 & 0 \\
Solar with Weather & energy & H & 24 & 20 & 49,648 & 1 & 1 & 2 & 7 & 0 \\
SZ Taxi & mobility & 15T & 96 & 10 & 2,976 & 156 & 1 & 0 & 0 & 0 \\
SZ Taxi & mobility & H & 168 & 2 & 744 & 156 & 1 & 0 & 0 & 0 \\
UCI Air Quality & nature & D & 28 & 11 & 389 & 1 & 4 & 0 & 3 & 0 \\
UCI Air Quality & nature & H & 168 & 20 & 9,357 & 1 & 4 & 0 & 3 & 0 \\
UK COVID - Nation - Cumulative & healthcare & D & 28 & 20 & 729 & 4 & 3 & 5 & 0 & 0 \\
UK COVID - Nation - Cumulative & healthcare & W & 8 & 4 & 105 & 4 & 3 & 5 & 0 & 0 \\
UK COVID - Nation - New & healthcare & D & 28 & 20 & 729 & 4 & 3 & 5 & 0 & 0 \\
UK COVID - Nation - New & healthcare & W & 8 & 4 & 105 & 4 & 3 & 5 & 0 & 0 \\
UK COVID - UTLA - Cumulative & healthcare & W & 13 & 5 & 104 & 214 & 1 & 0 & 0 & 0 \\
UK COVID - UTLA - New & healthcare & D & 28 & 10 & 721 & 214 & 1 & 0 & 0 & 0 \\
US Consumption & econ & M & 12 & 10 & 792 & 31 & 1 & 0 & 0 & 0 \\
US Consumption & econ & Q & 8 & 10 & 262 & 31 & 1 & 0 & 0 & 0 \\
US Consumption & econ & Y & 5 & 10 & 64 & 31 & 1 & 0 & 0 & 0 \\
Walmart & retail & W & 39 & 1 & 143 & 2,936 & 1 & 0 & 10 & 4 \\
World CO2 Emissions & econ & Y & 5 & 9 & 60 & 191 & 1 & 0 & 0 & 0 \\
World Life Expectancy & econ & Y & 5 & 10 & 74 & 237 & 1 & 0 & 0 & 0 \\
World Tourism & econ & Y & 5 & 2 & 21 & 178 & 1 & 0 & 0 & 0 \\
\bottomrule
\end{tabular}%

}
\caption{\textbf{Task statistics for \fevbench (Part 2/2).}}

\end{table}

\begin{table}
\centering
\resizebox{\textwidth}{!}{%

\begin{tabular}{lllrrrrr}
\toprule
\textbf{Task} & \textbf{Domain} & \textbf{Freq.} & $H$ & $W$ & \textbf{Avg. length} & \textbf{\# series} & \textbf{\# targets} \\
\midrule
Australian Electricity & energy & 30T & 48 & 1 & 231052 & 5 & 1 \\
CIF 2016 & banking & M & 12 & 1 & 98 & 72 & 1 \\
Car Parts & retail & M & 12 & 1 & 51 & 2674 & 1 \\
Covid Deaths & healthcare & D & 30 & 1 & 212 & 266 & 1 \\
Dominick & retail & D & 8 & 1 & 296 & 100014 & 1 \\
ERCOT Load & energy & H & 24 & 1 & 154854 & 8 & 1 \\
ETT (15 Min.) & energy & 15T & 24 & 1 & 69680 & 14 & 1 \\
ETT (Hourly) & energy & H & 24 & 1 & 17420 & 14 & 1 \\
Exchange Rate & finance & B & 30 & 1 & 7588 & 8 & 1 \\
FRED-MD & economics & M & 12 & 1 & 728 & 107 & 1 \\
Hospital & healthcare & M & 12 & 1 & 84 & 767 & 1 \\
M1 (Monthly) & various & M & 18 & 1 & 90 & 617 & 1 \\
M1 (Quarterly) & various & Q & 8 & 1 & 48 & 203 & 1 \\
M1 (Yearly) & various & Y & 6 & 1 & 24 & 181 & 1 \\
M3 (Monthly) & various & M & 18 & 1 & 117 & 1428 & 1 \\
M3 (Quarterly) & various & Q & 8 & 1 & 48 & 756 & 1 \\
M3 (Yearly) & various & Y & 6 & 1 & 28 & 645 & 1 \\
M4 (Quarterly) & various & Q & 8 & 1 & 100 & 24000 & 1 \\
M4 (Yearly) & various & Y & 6 & 1 & 37 & 23000 & 1 \\
M5 & retail & D & 28 & 1 & 1562 & 30490 & 1 \\
NN5 (Daily) & finance & D & 56 & 1 & 791 & 111 & 1 \\
NN5 (Weekly) & finance & W & 8 & 1 & 113 & 111 & 1 \\
Tourism (Monthly) & various & M & 24 & 1 & 298 & 366 & 1 \\
Tourism (Quarterly) & various & Q & 8 & 1 & 99 & 427 & 1 \\
Tourism (Yearly) & various & Y & 4 & 1 & 24 & 518 & 1 \\
Traffic & transport & H & 24 & 1 & 17544 & 862 & 1 \\
Weather & nature & D & 30 & 1 & 14296 & 3010 & 1 \\
\bottomrule
\end{tabular}%

}

\caption{\textbf{Task statistics for \texttt{Chronos-Bench-II}.} $H$ denotes the forecast horizon, $W$ the number of rolling evaluation windows.}
\label{tab:chronos-bench-ii-task-stats}

\end{table}

\begin{table}
\centering
\resizebox{\textwidth}{!}{%

\begin{tabular}{lllrrrrr}
\toprule
\textbf{Task} & \textbf{Domain} & \textbf{Freq.} & $H$ & $W$ & \textbf{Avg. length} & \textbf{\# series} & \textbf{\# targets} \\
\midrule
Jena Weather (10T, S) & Nature & 10T & 48 & 20 & 52,704 & 1 & 21 \\
Jena Weather (10T, M) & Nature & 10T & 480 & 11 & 52,704 & 1 & 21 \\
Jena Weather (10T, L) & Nature & 10T & 720 & 8 & 52,704 & 1 & 21 \\
Jena Weather (H, S) & Nature & H & 48 & 19 & 8,784 & 1 & 21 \\
Jena Weather (H, M) & Nature & H & 480 & 2 & 8,784 & 1 & 21 \\
Jena Weather (H, L) & Nature & H & 720 & 2 & 8,784 & 1 & 21 \\
Jena Weather (D, S) & Nature & D & 30 & 2 & 366 & 1 & 21 \\
BizITObs - Application (10S, S) & Web/CloudOps & 10S & 60 & 15 & 8,834 & 1 & 2 \\
BizITObs - Application (10S, M) & Web/CloudOps & 10S & 600 & 2 & 8,834 & 1 & 2 \\
BizITObs - Application (10S, L) & Web/CloudOps & 10S & 900 & 1 & 8,834 & 1 & 2 \\
BizITObs - Service (10S, S) & Web/CloudOps & 10S & 60 & 15 & 8,835 & 21 & 2 \\
BizITObs - Service (10S, M) & Web/CloudOps & 10S & 600 & 2 & 8,835 & 21 & 2 \\
BizITObs - Service (10S, L) & Web/CloudOps & 10S & 900 & 1 & 8,835 & 21 & 2 \\
BizITObs - L2C (5T, S) & Web/CloudOps & 5T & 48 & 20 & 31,968 & 1 & 7 \\
BizITObs - L2C (5T, M) & Web/CloudOps & 5T & 480 & 7 & 31,968 & 1 & 7 \\
BizITObs - L2C (5T, L) & Web/CloudOps & 5T & 720 & 5 & 31,968 & 1 & 7 \\
BizITObs - L2C (H, S) & Web/CloudOps & H & 48 & 6 & 2,664 & 1 & 7 \\
BizITObs - L2C (H, M) & Web/CloudOps & H & 480 & 1 & 2,664 & 1 & 7 \\
BizITObs - L2C (H, L) & Web/CloudOps & H & 720 & 1 & 2,664 & 1 & 7 \\
Bitbrains - Fast Storage (5T, S) & Web/CloudOps & 5T & 48 & 18 & 8,640 & 1,250 & 2 \\
Bitbrains - Fast Storage (5T, M) & Web/CloudOps & 5T & 480 & 2 & 8,640 & 1,250 & 2 \\
Bitbrains - Fast Storage (5T, L) & Web/CloudOps & 5T & 720 & 2 & 8,640 & 1,250 & 2 \\
Bitbrains - Fast Storage (H, S) & Web/CloudOps & H & 48 & 2 & 721 & 1,250 & 2 \\
Bitbrains - rnd (5T, S) & Web/CloudOps & 5T & 48 & 18 & 8,640 & 500 & 2 \\
Bitbrains - rnd (5T, M) & Web/CloudOps & 5T & 480 & 2 & 8,640 & 500 & 2 \\
Bitbrains - rnd (5T, L) & Web/CloudOps & 5T & 720 & 2 & 8,640 & 500 & 2 \\
Bitbrains - rnd (H, S) & Web/CloudOps & H & 48 & 2 & 720 & 500 & 2 \\
Restaurant (D, S) & Sales & D & 30 & 1 & 358 & 807 & 1 \\
ETT1 (15T, S) & Energy & 15T & 48 & 20 & 69,680 & 1 & 7 \\
ETT1 (15T, M) & Energy & 15T & 480 & 15 & 69,680 & 1 & 7 \\
ETT1 (15T, L) & Energy & 15T & 720 & 10 & 69,680 & 1 & 7 \\
ETT1 (H, S) & Energy & H & 48 & 20 & 17,420 & 1 & 7 \\
ETT1 (H, M) & Energy & H & 480 & 4 & 17,420 & 1 & 7 \\
ETT1 (H, L) & Energy & H & 720 & 3 & 17,420 & 1 & 7 \\
ETT1 (D, S) & Energy & D & 30 & 3 & 725 & 1 & 7 \\
ETT1 (W, S) & Energy & W & 8 & 2 & 103 & 1 & 7 \\
ETT2 (15T, S) & Energy & 15T & 48 & 20 & 69,680 & 1 & 7 \\
ETT2 (15T, M) & Energy & 15T & 480 & 15 & 69,680 & 1 & 7 \\
ETT2 (15T, L) & Energy & 15T & 720 & 10 & 69,680 & 1 & 7 \\
ETT2 (H, S) & Energy & H & 48 & 20 & 17,420 & 1 & 7 \\
ETT2 (H, M) & Energy & H & 480 & 4 & 17,420 & 1 & 7 \\
ETT2 (H, L) & Energy & H & 720 & 3 & 17,420 & 1 & 7 \\
ETT2 (D, S) & Energy & D & 30 & 3 & 725 & 1 & 7 \\
ETT2 (W, S) & Energy & W & 8 & 2 & 103 & 1 & 7 \\
Loop Seattle (5T, S) & Transport & 5T & 48 & 20 & 105,120 & 323 & 1 \\
Loop Seattle (5T, M) & Transport & 5T & 480 & 20 & 105,120 & 323 & 1 \\
Loop Seattle (5T, L) & Transport & 5T & 720 & 15 & 105,120 & 323 & 1 \\
Loop Seattle (H, S) & Transport & H & 48 & 19 & 8,760 & 323 & 1 \\
Loop Seattle (H, M) & Transport & H & 480 & 2 & 8,760 & 323 & 1 \\
\bottomrule
\end{tabular}%

}

\caption{\textbf{Task statistics for \texttt{GIFT-Eval} (Part 1/2).} $H$ denotes the forecast horizon, $W$ the number of rolling evaluation windows. S/M/L denote short/medium/long-term prediction horizons.}
\label{tab:gift-eval-task-stats}

\end{table}

\begin{table}
\centering
\resizebox{\textwidth}{!}{%

\begin{tabular}{lllrrrrr}
\toprule
\textbf{Task} & \textbf{Domain} & \textbf{Freq.} & $H$ & $W$ & \textbf{Avg. length} & \textbf{\# series} & \textbf{\# targets} \\
\midrule
Loop Seattle (H, L) & Transport & H & 720 & 2 & 8,760 & 323 & 1 \\
Loop Seattle (D, S) & Transport & D & 30 & 2 & 365 & 323 & 1 \\
SZ-Taxi (15T, S) & Transport & 15T & 48 & 7 & 2,976 & 156 & 1 \\
SZ-Taxi (15T, M) & Transport & 15T & 480 & 1 & 2,976 & 156 & 1 \\
SZ-Taxi (15T, L) & Transport & 15T & 720 & 1 & 2,976 & 156 & 1 \\
SZ-Taxi (H, S) & Transport & H & 48 & 2 & 744 & 156 & 1 \\
M\_DENSE (H, S) & Transport & H & 48 & 20 & 17,520 & 30 & 1 \\
M\_DENSE (H, M) & Transport & H & 480 & 4 & 17,520 & 30 & 1 \\
M\_DENSE (H, L) & Transport & H & 720 & 3 & 17,520 & 30 & 1 \\
M\_DENSE (D, S) & Transport & D & 30 & 3 & 730 & 30 & 1 \\
Solar (10T, S) & Energy & 10T & 48 & 20 & 52,560 & 137 & 1 \\
Solar (10T, M) & Energy & 10T & 480 & 11 & 52,560 & 137 & 1 \\
Solar (10T, L) & Energy & 10T & 720 & 8 & 52,560 & 137 & 1 \\
Solar (H, S) & Energy & H & 48 & 19 & 8,760 & 137 & 1 \\
Solar (H, M) & Energy & H & 480 & 2 & 8,760 & 137 & 1 \\
Solar (H, L) & Energy & H & 720 & 2 & 8,760 & 137 & 1 \\
Solar (D, S) & Energy & D & 30 & 2 & 365 & 137 & 1 \\
Solar (W, S) & Energy & W & 8 & 1 & 52 & 137 & 1 \\
Hierarchical Sales (D, S) & Sales & D & 30 & 7 & 1,825 & 118 & 1 \\
Hierarchical Sales (W, S) & Sales & W & 8 & 4 & 260 & 118 & 1 \\
M4 Yearly (Y, S) & Econ/Fin & Y & 6 & 1 & 37 & 22,974 & 1 \\
M4 Quarterly (Q, S) & Econ/Fin & Q & 8 & 1 & 100 & 24,000 & 1 \\
M4 Monthly (M, S) & Econ/Fin & M & 18 & 1 & 234 & 48,000 & 1 \\
M4 Weekly (W, S) & Econ/Fin & W & 13 & 1 & 1,035 & 359 & 1 \\
M4 Daily (D, S) & Econ/Fin & D & 14 & 1 & 2,371 & 4,227 & 1 \\
M4 Hourly (H, S) & Econ/Fin & H & 48 & 2 & 902 & 414 & 1 \\
Hospital (M, S) & Healthcare & M & 12 & 1 & 84 & 767 & 1 \\
COVID Deaths (D, S) & Healthcare & D & 30 & 1 & 212 & 266 & 1 \\
US Births (D, S) & Healthcare & D & 30 & 20 & 7,305 & 1 & 1 \\
US Births (W, S) & Healthcare & W & 8 & 14 & 1,043 & 1 & 1 \\
US Births (M, S) & Healthcare & M & 12 & 2 & 240 & 1 & 1 \\
Saugeen (D, S) & Nature & D & 30 & 20 & 23,741 & 1 & 1 \\
Saugeen (W, S) & Nature & W & 8 & 20 & 3,391 & 1 & 1 \\
Saugeen (M, S) & Nature & M & 12 & 7 & 780 & 1 & 1 \\
Temperature Rain (D, S) & Nature & D & 30 & 3 & 725 & 32,072 & 1 \\
KDD Cup 2018 (H, S) & Nature & H & 48 & 20 & 10,898 & 270 & 1 \\
KDD Cup 2018 (H, M) & Nature & H & 480 & 2 & 10,898 & 270 & 1 \\
KDD Cup 2018 (H, L) & Nature & H & 720 & 2 & 10,898 & 270 & 1 \\
KDD Cup 2018 (D, S) & Nature & D & 30 & 2 & 455 & 270 & 1 \\
Car Parts (M, S) & Sales & M & 12 & 1 & 51 & 2,674 & 1 \\
Electricity (15T, S) & Energy & 15T & 48 & 20 & 140,256 & 370 & 1 \\
Electricity (15T, M) & Energy & 15T & 480 & 20 & 140,256 & 370 & 1 \\
Electricity (15T, L) & Energy & 15T & 720 & 20 & 140,256 & 370 & 1 \\
Electricity (H, S) & Energy & H & 48 & 20 & 35,064 & 370 & 1 \\
Electricity (H, M) & Energy & H & 480 & 8 & 35,064 & 370 & 1 \\
Electricity (H, L) & Energy & H & 720 & 5 & 35,064 & 370 & 1 \\
Electricity (D, S) & Energy & D & 30 & 5 & 1,461 & 370 & 1 \\
Electricity (W, S) & Energy & W & 8 & 3 & 208 & 370 & 1 \\
\bottomrule
\end{tabular}%

}
\caption{\textbf{Task statistics for \texttt{GIFT-Eval} (Part 2/2).}}
\end{table}

\end{document}